\documentclass[11pt]{article}

\usepackage[final]{acl}

\usepackage{times}
\usepackage{latexsym}
\usepackage{wrapfig}

\usepackage[T1]{fontenc}

\usepackage[utf8]{inputenc}

\usepackage{microtype}

\usepackage{inconsolata}

\usepackage{multirow}
\usepackage{graphicx}     
\usepackage{subcaption}
\usepackage{enumitem}
\usepackage{amsmath}
\usepackage{amsthm}
\usepackage{algorithm}
\usepackage{algorithmic}  
\usepackage[table]{xcolor}
\usepackage{svg}
\usepackage{makecell}
\usepackage{pifont}
\usepackage{enumitem}
\usepackage{amssymb}  
\usepackage{tablefootnote}
\usepackage{booktabs}

\newtheorem{theorem}{Theorem}
\newtheorem{definition}{Definition}

\title{VisPCO: Visual Token Pruning Configuration Optimization via Budget-Aware Pareto-Frontier Learning for Vision-Language Models}

\author{
 \textbf{Huawei Ji\textsuperscript{1}},
 \textbf{Yuanhao Sun\textsuperscript{1}},
 \textbf{Yuan Jin\textsuperscript{1}},
 \textbf{Cheng Deng\textsuperscript{2}},
\\
    \textbf{Jiaxin Ding\textsuperscript{1}\thanks{Corresponding author}}
 \textbf{Luoyi Fu\textsuperscript{1}},
 \textbf{Xinbing Wang\textsuperscript{1}}
\\
\\
\textsuperscript{1}Shanghai Jiao Tong University, Shanghai, China,\\
\textsuperscript{2}University of Edinburgh, Edinburgh, UK
\\
  \texttt{\{sjtu3365981, h\_iden, lemon0703, jiaxinding, yiluofu, xwang8\}}@sjtu.edu.cn \\}

\begin{document}
\maketitle
\begin{abstract}
Visual token pruning methods effectively mitigate the quadratic computational growth caused by processing high-resolution images or long video frames in vision-language models (VLMs). However, existing approaches rely on predefined pruning configurations without determining whether they achieve computation-performance optimality. In this work, we introduce \textbf{VisPCO}, a novel framework that formulates visual token pruning as a Pareto configuration optimization problem to automatically identify optimal configurations. Our approach employs continuous relaxation and straight-through estimators to enable gradient-based search, solved via the Augmented Lagrangian method. Extensive experiments across 8 visual benchmarks demonstrate that \textbf{VisPCO} effectively approximates the empirical Pareto frontier obtained through grid search and generalizes well across various pruning methods and VLM architectures. Furthermore, through learnable kernel functions, we investigate layer-wise pruning patterns and reveal that multi-step progressive pruning captures VLMs' hierarchical compression structure, achieving superior computation-performance trade-offs compared to single-layer approaches. 
\end{abstract}

\section{Introduction}
Large-scale vision-language models (LVLMs) process both visual and textual features as input, enabling them to learn unified multimodal representations and perform cross-modal reasoning. Recent studies have shown that higher-resolution image inputs can effectively improve the model's understanding and generation performance~\citep{LLaVA-UHD,LLaVAOneVision,InternVL25}. Meanwhile, tasks such as video understanding require models to process numerous frames to capture temporal continuity and dynamic semantics~\citep{VideoLLaVA, StreamingVLM}. Both scenarios significantly increase the number of visual tokens, leading to quadratic growth in computational costs.

\begin{figure}[t]
    \centering
    \captionsetup[subfigure]{justification=centering, skip=1pt}
    
    \begin{subfigure}[b]{0.50\columnwidth}
        \centering
        \includegraphics[width=\textwidth, height=3cm]{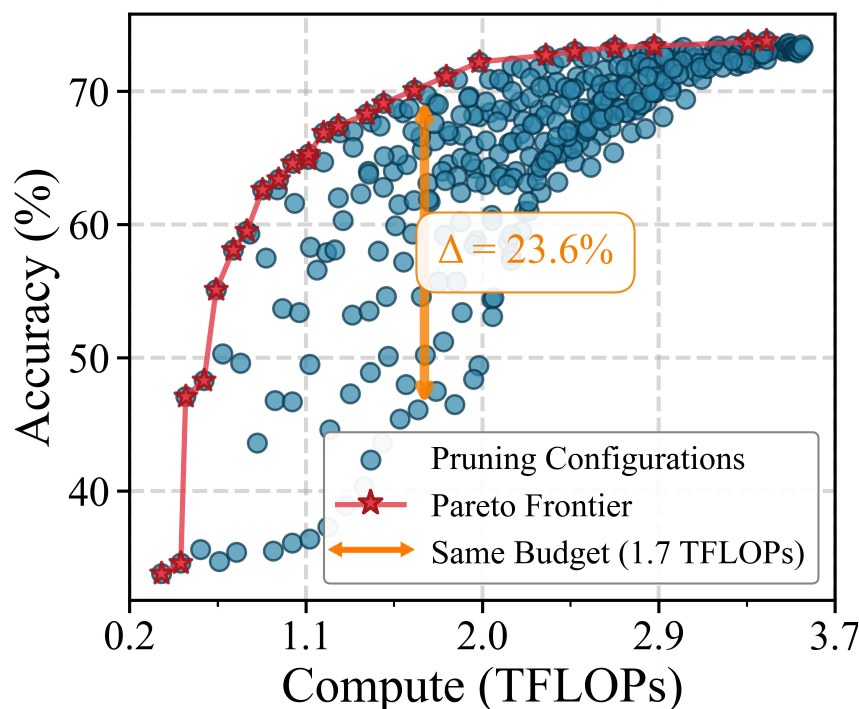}
        \caption{Pareto Frontier}
        \label{fig:1a}
    \end{subfigure}%
    \hfill
    \begin{subfigure}[b]{0.50\columnwidth}
        \centering
        \includegraphics[width=\textwidth, height=3cm]{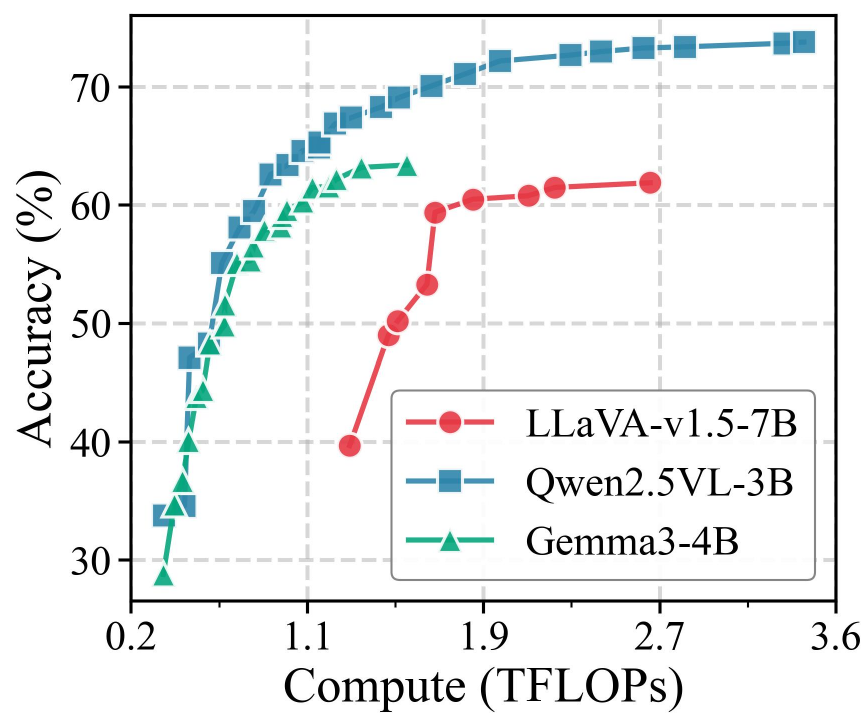}
        \caption{Base Models}
        \label{fig:1b}
    \end{subfigure}
    
    
    \begin{subfigure}[b]{0.50\columnwidth}
        \centering
        \includegraphics[width=\textwidth, height=3cm]{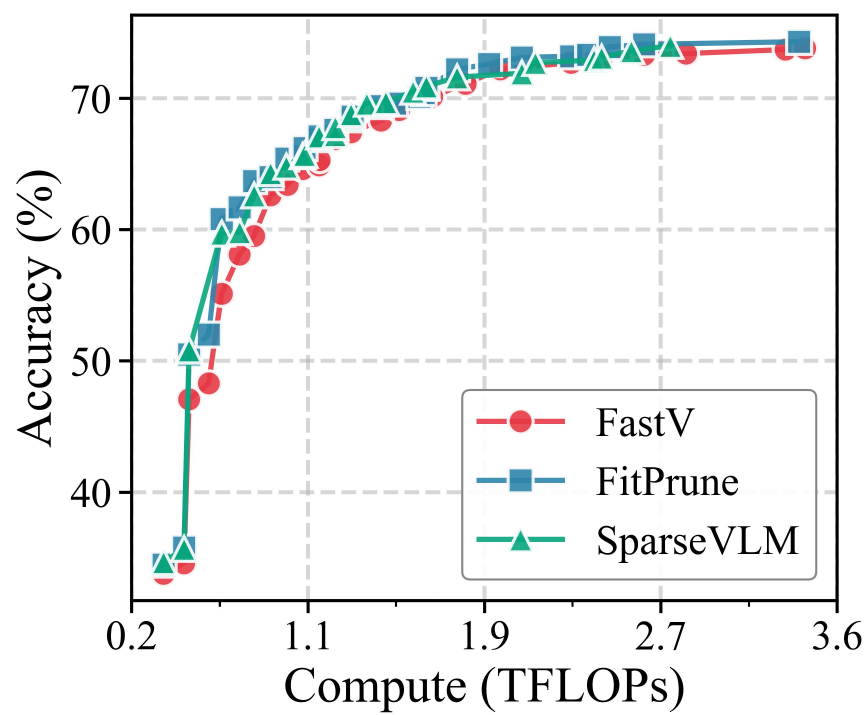}
        \caption{Pruning Methods}
        \label{fig:1c}
    \end{subfigure}%
    \hfill
    \begin{subfigure}[b]{0.50\columnwidth}
        \centering
        \includegraphics[width=\textwidth, height=3cm]{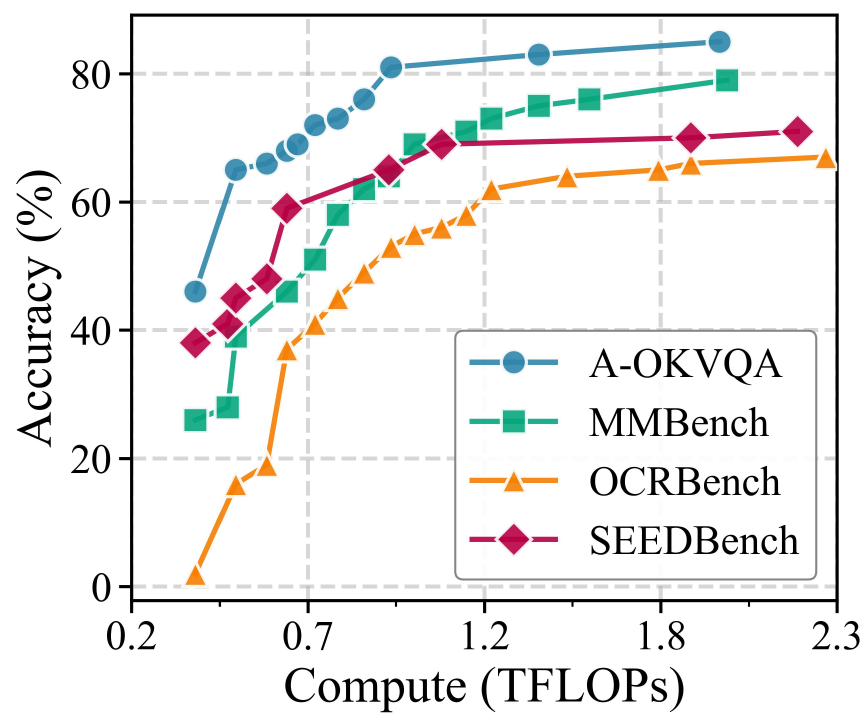}
        \caption{Different Datasets}
        \label{fig:1d}
    \end{subfigure}
    
    \caption{Pareto frontiers across different configurations: (a) The Pareto frontier connects optimal pruning configurations. (b) Pareto frontiers across different VLMs. (c) Pareto frontiers across different pruning methods. (d) Pareto frontiers across different datasets.}

    \label{fig:pareto_curve}
    \vspace{-0.5cm}
\end{figure}

To solve this problem, various visual pruning algorithms for VLMs have emerged. These methods mainly focus on designing different importance scoring mechanisms to prune redundant visual tokens at single or multiple layers. For instance, FastV~\citep{FastV}, Dynamic-LLaVA~\citep{Dynamic-LLaVA}, VTW~\citep{VTW}, and TOPV~\citep{TOPV} prune visual tokens at a specific LLM layer using predefined pruning ratios. In contrast, ATP-LLaVA~\citep{APT-LLaVA}, HiMAP~\citep{HiMAP}, and SparseVLM~\citep{SparseVLM} apply dynamic pruning ratios across multiple selected layers. All these works aim to reduce computational costs (e.g., FLOPs) while maintaining model performance.

\begin{figure*}[!t]
    \centering
    \includegraphics[width=\textwidth]{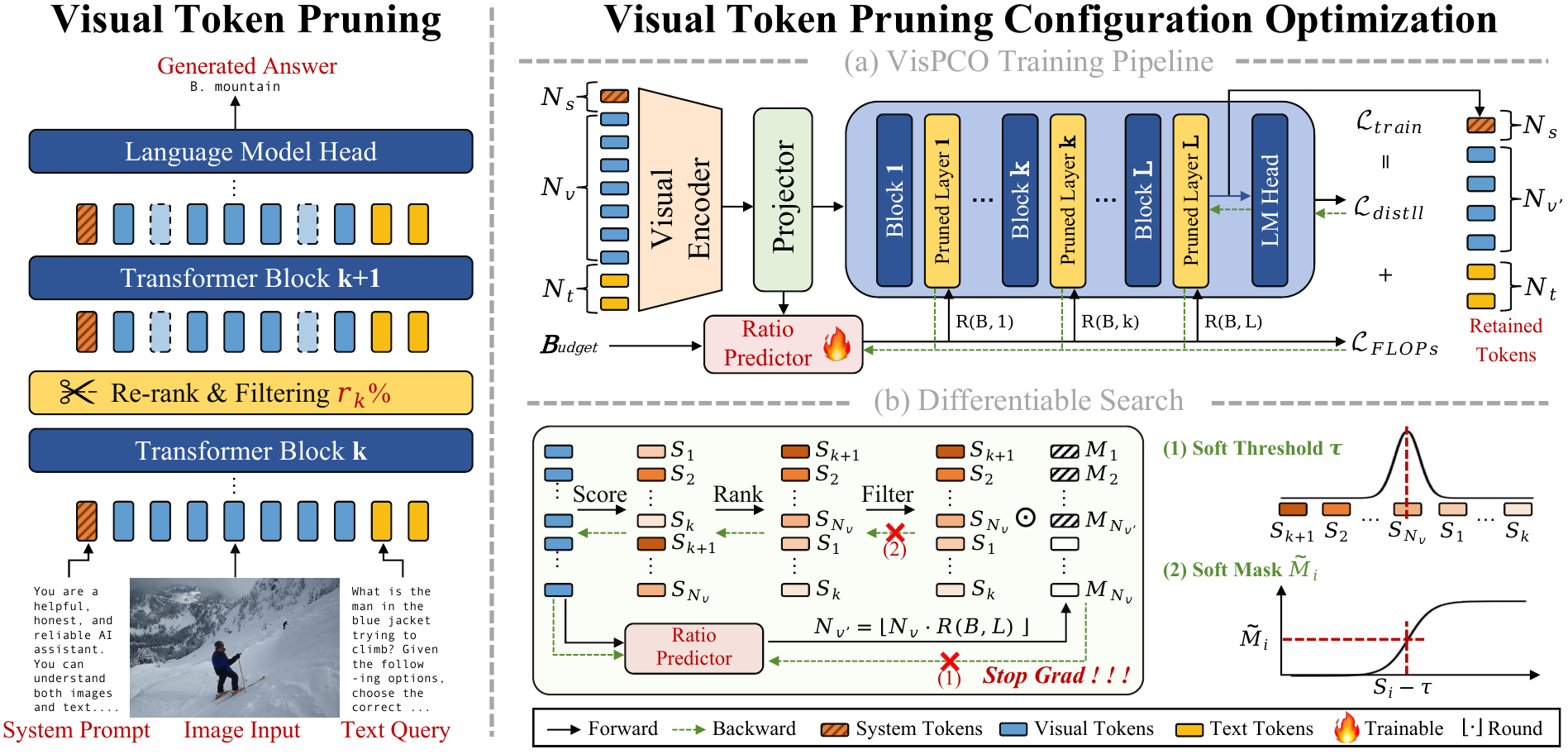}
    \caption{Illustration of our \textbf{VisPCO} framework. \textbf{(Left)} Overview of the visual token pruning process. After each transformer block, visual tokens are ranked by their importance scores and low-scoring tokens are filtered out. \textbf{(Right)} Upper panel: The overall architecture of \textbf{VisPCO}, where the trainable Ratio Predictor, a lightweight surrogate network, determines the pruning ratio to guide token compression at each layer. Lower panel: The gradient disconnection problem encountered during end-to-end training of the Ratio Predictor and our proposed solution.}
    \label{fig:architecture}
    \vspace{-0.5cm}
\end{figure*}

However, two critical questions remain unexplored in existing works. First, it is unclear whether current pruning configurations (i.e., pruning positions and ratios) achieve the optimal computation-performance trade-off. Second, how to adjust these configurations efficiently to reach the optimal remains an open problem. Addressing these two questions is crucial for the efficient deployment of VLMs in real-world applications, e.g., edge device deployment and mobile vision systems. 

To this end, we introduce the concept of Pareto optimality to characterize the optimal computation-performance trade-off~\citep{ParetoOptimization}. Pareto optimality is a classic concept in multi-objective optimization that describes a state where no further improvement can be made among multiple conflicting objectives. In this work, we define a point as Pareto-optimal if we cannot simultaneously reduce computational cost and improve performance. The curve connecting these points is called the Pareto frontier. 

In Figure~\ref{fig:pareto_curve}, we visualize different pruning configurations in the computation-performance space and their corresponding Pareto frontier. Each point represents the experimental result of one configuration. Figure~\ref{fig:pareto_curve}(a) reveals that under the same computational budget, different pruning configurations exhibit significant performance variations. For instance, at a computational budget of 1.7 TFLOPs, the performance gap between different configurations can reach up to 23.6\%. Moreover, Pareto-optimal frontiers are not fixed. Instead, they vary with VLM architectures, pruning methods and image complexity as illustrated in Figure~\ref{fig:pareto_curve}(b-d). In practice, to determine the best pruning configuration under a fixed computational budget, we typically need to perform grid sampling over the configuration search space. This involves conducting numerous experiments to measure performance and cost across different configurations, and then selecting the optimal one for deployment. However, this process is prohibitively time-consuming and resource-intensive.

In this paper, we propose a computation budget-aware method for \textbf{Vis}ual token \textbf{P}runing \textbf{C}onfiguration \textbf{O}ptimization, termed \textbf{VisPCO}. This approach employs a learnable surrogate model to automatically predict pruning configurations on the Pareto frontier given a computational budget, thereby achieving optimal model performance. As shown in Figure~\ref{fig:architecture}, unlike traditional grid search methods, \textbf{VisPCO} employs efficient gradient descent for search, significantly reducing search costs. To address the discrete and non-differentiable nature of visual token pruning, we introduce continuous relaxation techniques and straight-through estimators for end-to-end optimization. For the optimization objective, we formulate it as a Pareto optimization problem with non-convex inequality constraints and solve it using the Augmented Lagrangian method~\citep{Aug-Lang}. Furthermore, we investigate the layer-wise pruning patterns in VLMs. Specifically, we explore how the optimal pruning ratio varies across different layers when progressively compressing visual tokens. To model this variation, we use learnable kernel functions to parameterize the pruning ratio distribution across layers. By evaluating how well different kernel functions approximate the Pareto frontier, we identify the intrinsic pruning patterns that lead to optimal computation-performance trade-offs. Our contributions can be summarized as follows:

\begin{itemize}[leftmargin=*, itemsep=0pt]
\vspace{-5pt}
    \item We propose \textbf{VisPCO}, a differentiable framework that automatically finds Pareto-optimal configurations via gradient-based methods, eliminating the prohibitive cost of exhaustive grid search.
    \item Experiments on 8 benchmarks demonstrate that \textbf{VisPCO} effectively approximates the empirical Pareto frontier and generalizes across various pruning methods and VLM architectures.
    \item We reveal non-uniform visual token redundancy across layers via learnable pruning kernels, showing multi-step pruning is most effective under tight budgets. Our code is available at \url{https://github.com/JHW5981/VisPCO}.
\end{itemize}




\section{Related Work}
\subsection{Visual Token Pruning}
Visual token pruning accelerates VLMs by reducing computational costs from processing hundreds of visual tokens. Single-layer approaches perform one-shot reduction at specific layers: FastV~\citep{FastV} prunes tokens after layer 2 using attention scores, Dynamic-LLaVA~\citep{Dynamic-LLaVA} dynamically adjusts token retention ratios based on input characteristics, VTW~\citep{VTW} withdraws all tokens after sufficient absorption, and TopV~\citep{TOPV} optimizes configurations at inference time. Multi-layer progressive methods distribute reduction across multiple layers: ATP-LLaVA~\citep{APT-LLaVA} adaptively prunes tokens at different depths with layer-specific ratios, SparseVLM~\citep{SparseVLM} progressively reduces tokens across layers with recycling mechanisms, and PyramidDrop~\citep{PyramidDrop} implements stage-wise pyramid reduction that preserves more tokens in shallow layers. These strategies achieve 40-90\% computational savings while maintaining competitive performance. However, the pruning configurations in these methods are either predefined or heuristically determined, and it remains unclear whether they achieve the optimal computation-performance trade-off.

\subsection{Pruning Configuration Optimization}
While most pruning methods rely on predefined configurations, recent works explore adaptive strategies to optimize pruning ratios across layers. FitPrune~\citep{FitPrune} employs binary search over attention statistics to minimize distribution divergence and generate layer-wise pruning recipes. G-Search~\citep{G-Search} combine greedy search with Bayesian-optimized sigmoid functions to approximate optimal retention ratios. ATP-LLaVA~\citep{APT-LLaVA} introduces learnable modules for training-based optimization of layer-specific sparsity. SparseVLM~\citep{SparseVLM} employs rank-based adaptive determination of per-layer sparsification ratios. More recent methods explore input-adaptive configurations: AIM~\citep{AIM} develops scheduler-controlled pruning with adjustable parameters, and MADTP~\citep{MADTP} utilizes learnable thresholds for instance-wise adaptive pruning. Despite these advances, existing approaches focus primarily on performance preservation rather than budget-aware optimization. They lack mechanisms to systematically identify near-optimal configurations under varying computational constraints.

\section{VisPCO}

\subsection{Pareto Optimization}
\label{sec:optimization objective}
We formulate the visual pruning configuration optimization problem as finding the optimal layer-wise pruning ratios $\mathbf{r} = [r_1, r_2, \ldots, r_L] \in [0,1]^L$, where $L$ denotes the number of layers and $r_i$ represents the token retention ratio at layer $i$ (relative to the original number of visual tokens). Our objective is to identify a configuration $\bar{\mathbf{r}}$ on the Pareto frontier that achieves optimal model performance under a given computational budget. 

To quantify performance degradation, we define the pruned VLM output logits as $\hat{{l}}$ and the original output logits as ${l}$, and measure their discrepancy using KL divergence:

\vspace{-1em}
\begin{equation}
\mathcal{L}_{\text{distill}}(\mathbf{r}) 
= D_{KL}\!\left( 
\mathrm{softmax}(\hat{{l}}) \,\middle\|\, 
\mathrm{softmax}({l})
\right).
\end{equation}

A smaller value of $\mathcal{L}_{\text{distill}}$ indicates less impact of pruning on model performance. Meanwhile, we define the computational cost function as:

\vspace{-1em}
\begin{small}
\begin{equation}
F(\mathbf{r}) = \sum_{i=1}^{L}\left[24(N_t + r_iN_v)D^2 + 4(N_t + r_iN_v)^2D\right]
\end{equation}
\end{small}
where $N_t$ is the number of text tokens, $N_v$ is the number of visual tokens, and $D$ is the hidden dimension. Detailed derivation of the FLOPs computation is provided in Appendix~\ref{Appendix: FLOPs Analysis}. Therefore, the Pareto optimization problem for the computation-performance trade-off can be formulated as the following constrained optimization problem:

\begin{equation}
\label{eq:objective}
\begin{aligned}
    \min \quad & \mathcal{L}_{\text{distill}}(\mathbf{r}) \\
    \text{s.t.} \quad & F(\mathbf{r}) \leq B,
\end{aligned}
\end{equation}
where $B$ is the given computational budget. Considering that the objective function $\mathcal{L}_{\text{distill}}$ is typically non-convex, we employ the Augmented Lagrangian Method for numerical iterative solving.

\begin{definition}[Augmented Lagrangian Method]
\label{def:augmented_lagrangian_method}
Consider the equality-constrained optimization problem:
\begin{equation*}
\begin{aligned}
\min \quad & f(\mathbf{x}) \\
\text{s.t.} \quad h_j(\mathbf{x}) &= 0, \quad j = 1,\ldots,l
\end{aligned}
\end{equation*}
where $f(\mathbf{x}): \mathbb{R}^n \rightarrow \mathbb{R}$ is the objective function to be minimized, and $h_j(\mathbf{x}): \mathbb{R}^n \rightarrow \mathbb{R}$ are the equality constraint functions. The augmented Lagrangian function is defined as:

\begin{small}
\vspace{-1em}
\begin{equation}
\label{eq:augmented_lagrangian_function}
\phi(\mathbf{x}, \mathbf{v}, \lambda) = f(\mathbf{x}) - \sum_{j=1}^l v_j h_j(\mathbf{x}) + \frac{\lambda}{2}\sum_{j=1}^l h_j^2(\mathbf{x}),
\end{equation}
\end{small}
where $\mathbf{v} = [v_1, \ldots, v_l]$ is the Lagrange multiplier vector and $\lambda > 0$ is the penalty parameter.
\end{definition}

\begin{theorem}[Adapted from~\citealp{Lang-Hard}]
\label{thm:convergence}
Let $\bar{\mathbf{x}}$ and $\bar{\mathbf{v}}$ satisfy the second-order conditions for a local optimal solution of the problem. Then there exists $\lambda' \geq 0$ such that for all $\lambda > \lambda'$, $\bar{\mathbf{x}}$ is a strict local minimizer of $\phi(\mathbf{x}, \bar{\mathbf{v}}, \lambda)$. 
\end{theorem}

The proof of Theorem~\ref{thm:convergence} is provided in Appendix~\ref{Appendix:proof_theorem1}. Based on this theorem, we can develop an iterative algorithm with bounded $\lambda$, avoiding the ill-conditioning of quadratic penalty methods~\citep{Aug-Lang} and the convergence difficulties of standard Lagrangian methods~\citep{Lang-Hard}. Specifically, at iteration $k$, let $\mathbf{x}^{(k)}$ be the minimizer of~\eqref{eq:augmented_lagrangian_function} with respect to $\mathbf{x}$. The multiplier update rule is:

\vspace{-1em}
\begin{equation}
v_j^{(k+1)} = v_j^{(k)} - \lambda h_j(\mathbf{x}^{(k)}), \quad j=1,\ldots,l
\end{equation}

Through this iterative update, we have $\mathbf{v}^{(k)} \to \bar{\mathbf{v}}$ and $\mathbf{x}^{(k)} \to \bar{\mathbf{x}}$, with convergence rate typically measured by $\|h(\mathbf{x}^{(k)})\| / \|h(\mathbf{x}^{(k-1)})\|$.

Returning to our optimization problem~\eqref{eq:objective}, we introduce an auxiliary variable $y$ to convert the inequality constraint into an equality constraint:

\begin{equation}
\label{eq:equality}
\begin{aligned}
\min \quad & \mathcal{L}_{\text{distill}}(\mathbf{r}) \\
\text{s.t.} \quad &B - F(\mathbf{r}) - y^2 = 0.
\end{aligned}
\end{equation}

The corresponding augmented Lagrangian function is defined as:

\begin{small}
\vspace{-1em}
\begin{multline}
\label{eq:our_augmented_lagrangian}
\tilde{\phi}(\mathbf{r}, y, w, \lambda) = \mathcal{L}_{\text{distill}}(\mathbf{r}) - w\left(B - F(\mathbf{r}) - y^2\right) \\
+ \frac{\lambda}{2}\left(B - F(\mathbf{r}) - y^2\right)^2.
\end{multline}
\end{small}

By completing the square with respect to \(y\), we can eliminate the dependence on \(y\) and obtain the simplified augmented Lagrangian (see Appendix~\ref{Appendix: eliminate y}):

\vspace{-1em}
\begin{equation}
\label{eq:simplified_lagrangian}
\begin{aligned}
\phi(\mathbf{r}, w, \lambda) 
=\, \mathcal{L}_{\text{distill}}(\mathbf{r})
+ \frac{1}{2\lambda}\left( z^2 - w^2 \right),
\end{aligned}
\end{equation}
where $z = \max\!\left(0,\; w - \lambda \bigl(B - F(\mathbf{r})\bigr) \right)$. The problem is thus transformed into minimizing the unconstrained objective $\phi(\mathbf{r}, w, \lambda)$. Using the iterative algorithm in Algorithm~\ref{alg:augmented_lagrangian}, we can obtain the optimal pruning configuration $\bar{\mathbf{r}}$. 

\subsection{Differentiable Configuration Search}
\label{sec: Relaxation}
Despite having the Pareto optimization objective Eq.~\eqref{eq:simplified_lagrangian} and iterative Algorithm~\ref{alg:augmented_lagrangian}, computing $\nabla_{\mathbf{r}} \mathcal{L}_{\text{distill}}(\mathbf{r})$ faces two critical non-differentiability challenges, as illustrated in the bottom right of Figure~\ref{fig:architecture}. First, the discretization of retained token counts introduces non-differentiability. For layer $i$ with pruning ratio $r_i$ and $N_v$ visual tokens, the retained token count $k_i = \lfloor r_i \cdot N_v \rfloor$ involves a floor operation that causes vanishing gradients, preventing backpropagation-based updates of $r_i$. Second, selecting the top-$k_i$ tokens based on importance scores involves discrete operations that block gradient flow. We propose the following two methods to address these challenges respectively:

\textbf{Continuous Relaxation.} 
To address the first challenge, we adopt a continuous relaxation strategy. We retain the floating-point form $\tilde{k}_i = r_i \cdot N_v$ and design a soft interpolation method using a Gaussian kernel to estimate the selection threshold. Specifically, we first sort the importance scores of all visual tokens in descending order to obtain $\{s_{i1}, s_{i2}, \ldots, s_{iN_v}\}$, where $s_{ij}$ denotes the score of the $j$-th token in layer $i$ after sorting. Traditional hard thresholding directly uses the score at position $\lfloor \tilde{k}_i \rfloor$ as the threshold, which leads to vanishing gradients. Instead, we employ Gaussian kernel-based soft interpolation to maintain differentiability:

\begin{small}
\vspace{-1em}
\begin{equation}
w_{ij} = \exp\left(-\frac{(j - \tilde{k}_i)^2}{2\sigma^2}\right), \quad 
\tau_i = \frac{\sum_{j=1}^{N_v} w_{ij} s_{ij}}{\sum_{j=1}^{N_v} w_{ij}},
\label{eq:soft_threshold}
\end{equation}
\end{small}

where $w_{ij}$ is the Gaussian weight for the $j$-th token in layer $i$. The parameter $\sigma$ controls the kernel width, balancing between approximation accuracy and gradient stability. As $\sigma \to 0$, the soft threshold $\tau_i$ converges to the hard threshold $s_{i\lfloor \tilde{k}_i \rfloor}$. The gradient of $\tau_i$ with respect to $r_i$ can be expressed as:
\begin{equation}
\frac{\partial \tau_i}{\partial r_i} = N_v \sum_{j=1}^{N_v} \frac{w_{ij}(j - \tilde{k}_i)}{\sigma^2 \sum_{l=1}^{N_v} w_{il}} (s_{ij} - \tau_i),
\label{eq:gradient_tau}
\end{equation}
which remains well-defined for all $r_i \in (0, 1)$, enabling smooth gradient flow via backpropagation.

\textbf{Straight-Through Estimator.} 
To address the second challenge, we employ the Straight-Through Estimator (STE) strategy. Inspired by Gumbel-Softmax~\citep{gumbel_softmax}, we use discrete hard decisions in the forward pass and continuous soft approximations in the backward pass. Given the threshold $\tau_i$, for each visual token $j$ in layer $i$, we compute both a hard selection mask and a soft mask:
\begin{equation}
m_{ij} = \mathbb{I}[s_{ij} \geq \tau_i], \quad 
\tilde{m}_{ij} = \text{sigm}\left(\frac{s_{ij} - \tau_i}{T}\right),
\label{eq:masks}
\end{equation}
where $\mathbb{I}[\cdot]$ is the indicator function, $\text{sigm}(\cdot)$ is the sigmoid function, and $T$ is the temperature parameter. The soft mask $\tilde{m}_{ij}$ provides a smooth approximation: as $T \to 0^+$, $\tilde{m}_{ij} \to m_{ij}$. The STE combines both masks through:
\begin{equation}
\hat{m}_{ij} = \tilde{m}_{ij} + \text{sg}(m_{ij} - \tilde{m}_{ij}),
\label{eq:ste}
\end{equation}
where $\text{sg}(\cdot)$ denotes the stop-gradient operation. This formulation ensures $\hat{m}_{ij} = m_{ij}$ in the forward pass, while the backward gradient satisfies:

\begin{footnotesize}
\begin{equation}
\frac{\partial \mathcal{L}}{\partial \tau_i} = \sum_{j} \frac{\partial \mathcal{L}}{\partial \hat{m}_{ij}} \cdot \frac{\partial \tilde{m}_{ij}}{\partial \tau_i}
= -\sum_{j} \frac{\partial \mathcal{L}}{\partial \hat{m}_{ij}} \cdot \frac{\tilde{m}_{ij}(1-\tilde{m}_{ij})}{T},
\label{eq:ste_gradient}
\end{equation}
\end{footnotesize}
where $\frac{\partial \mathcal{L}}{\partial \hat{m}_{ij}}$ denotes the upstream gradient propagated from subsequent layers and is well-defined since $\hat{m}_{ij}$ is treated as a differentiable proxy of $m_{ij}$ during backpropagation. These provide biased but low-variance gradient estimators that enable end-to-end optimization of both token scores and adaptive thresholds.

\subsection{Learnable Kernel Functions}
\label{sec: Kernel Function}

With \textbf{VisPCO}, we can automatically search for configurations on the Pareto frontier in a differentiable manner. To further investigate how different pruning patterns affect the Pareto frontier, we impose structural constraints on the pruning configuration search space. In practice, visual token pruning exhibits a monotonically non-increasing pattern across layers: deeper layers tend to retain fewer tokens. We leverage this prior by introducing \textbf{learnable kernel functions} to parameterize the layer-wise pruning ratios. This design offers two key benefits: (1) it provides interpretability for visual token pruning behavior, and (2) it reduces the parameter search space, ensuring both optimization stability and computational efficiency.

\begin{algorithm}[!t]
   \caption{Training pipeline of \textbf{VisPCO}}\label{alg:augmented_lagrangian}
\begin{algorithmic}[1]
   \STATE {\bfseries Input:} Initial configuration $\mathbf{r}^{(0)}$, initial Lagrange multiplier $w^{(1)}$, penalty parameter $\lambda$, convergence threshold $\epsilon > 0$, update coefficients $\alpha > 1$, $\beta \in (0,1)$
   \STATE {\bfseries Output:} Locally optimal configuration $\bar{\mathbf{r}}$ and optimal multiplier $\bar{w}$
   \STATE Set $k=1$
   \WHILE{True}
   \STATE Starting from $\mathbf{r}^{(k-1)}$, solve optimization problem $\min \phi(\mathbf{r}, w, \lambda)$
   \STATE \textcolor{gray}{\textbf{\textit{/* Train using gradient descent */}}}
   \STATE Obtain solution $\mathbf{r}^{(k)}$
   \IF{$\|B - F(\mathbf{r}^{(k)})\| < \epsilon$}
   \STATE \textcolor{gray}{\textbf{\textit{/* Constraint satisfied, training converged */}}}
   \STATE \textbf{break}
   \ENDIF
   \IF{$\frac{\|B - F(\mathbf{r}^{(k)})\|}{\|B - F(\mathbf{r}^{(k-1)})\|} \geq \beta$}
   \STATE \textcolor{gray}{\textbf{\textit{/* Update penalty parameter */}}}
   \STATE $\lambda \gets \alpha \lambda$
   \ENDIF
   \STATE \textcolor{gray}{\textbf{\textit{/* Update Lagrange multiplier */}}}
   \STATE $w^{(k+1)} \gets w^{(k)} - \lambda (B - F(\mathbf{r}^{(k)}))$
   \STATE $k \gets k + 1$
   \ENDWHILE
   \STATE \textbf{return} $\bar{\mathbf{r}} = \mathbf{r}^{(k)}$, $\bar{w} = w^{(k)}$
\end{algorithmic}
\end{algorithm}

We consider two pruning scenarios: single-layer pruning and multi-layer pruning. For the first case, we employ a parameterized p-sigmoid kernel to model a sharp transition at layer $k$:

\begin{small}
\begin{equation}
\mathcal{K}_{\text{s}}(i; k, r, \gamma) = 1 + (r - 1) \cdot \text{sigm}(\gamma(i - k)),
\label{eq:kernel_single}
\end{equation}
\end{small}
where $i$ is the layer index, $k$ is the pruning position, $r \in (0, 1]$ controls the final retention ratio, $\gamma > 0$ is the sharpness parameter, and $\text{sigm}(\cdot)$ is the sigmoid function. With sufficiently large $\gamma$, the transition approaches a step function retaining all tokens before layer $k$ and applying retention ratio $r$ thereafter, effectively approximating single-layer pruning while remaining differentiable.

\begin{table*}[!t]
\centering
\caption{Comparison of pruning methods with and without \textbf{VisPCO} on eight benchmarks under different budgets.  Results without \textbf{VisPCO} are averaged over multiple sampled configurations meeting the budget constraint ($\pm$ std).}
\label{tab: exp1}
\resizebox{\textwidth}{!}{

\begin{tabular}{l|cccccccc|c}
\noalign{
  \hrule height 1pt
}
\textbf{Method} & \textbf{AOKVQA} & \textbf{VizWiz} & \textbf{SEED} & \textbf{MMB} & \textbf{MME}$^{\dagger}$ & \textbf{ChartQA} & \textbf{OCRB} & \textbf{TextVQA} & \textbf{Avg (\%)}\\
\noalign{
  \hrule height 1pt
}
\rowcolor{gray!20}
\multicolumn{10}{c}{\textit{Upper Bound, 100\% Budget, $\sim$3.56 TFLOPs}} \\
Qwen2.5VL-3B & $90.2$ & $75.1$ & $75.6$ & $79.8$ & $84.2$ & $64.1$ & $74.6$ & $81.3$ & $78.1$ \\
\noalign{
  \hrule height 1pt
}
\rowcolor{gray!20}
\multicolumn{10}{c}{\textit{Reduce FLOPs Budget to 90\%, $\sim$3.20 TFLOPs}} \\
$\llcorner$ FastV & $88.2 \pm 0.4$ & $72.9 \pm 0.9$ & $72.4 \pm 0.9$ & $76.4 \pm 0.5$ & $81.3 \pm 0.5$ & $62.2 \pm 0.8$ & $71.6 \pm 0.7$ & $79.1 \pm 0.6$ & $75.5 \pm 0.7$\\
\;\;+ VisPCO & $88.4$ & $73.8$ & $73.2$ & $76.9$ & $81.7$ & $62.9$ & $72.3$ & $79.5$ & $76.1$\\
$\llcorner$ SparseVLM & $88.5 \pm 0.3$ & $73.1 \pm 0.5$ & $73.4 \pm 0.4$ & $76.9 \pm 0.6$ & $82.1 \pm 0.3$ & $62.2 \pm 0.7$ & $71.9 \pm 0.6$ & $79.5 \pm 0.6$ & $76.0 \pm 0.5$\\
\;\;+ VisPCO & $88.6$ & $73.5$ & $73.8$ & $77.5$ & $82.4$ & $62.9$ & $72.5$ & $80.0$ & $76.4$\\
$\llcorner$ FitPrune & $89.1 \pm 0.5$ & $73.9 \pm 0.4$ & $74.2 \pm 0.5$ & $77.6 \pm 0.4$ & $82.5 \pm 0.6$ & $63.1 \pm 0.6$ & $72.5 \pm 0.5$ & $79.9 \pm 0.3$ & $76.2 \pm 0.5$\\
\;+\; VisPCO & $89.6$ & $74.1$ & $74.6$ & $77.9$ & $82.8$ & $63.5$ & $72.9$ & $81.2$ & $77.1$\\

\noalign{
  \hrule height 1pt
}
\rowcolor{gray!20}
\multicolumn{10}{c}{\textit{Reduce FLOPs Budget to 50\%, $\sim$3.56 TFLOPs}} \\
$\llcorner$ FastV& $74.7 \pm 10.1$ & $60.3 \pm 9.6$ & $61.5 \pm 8.1$ & $62.4 \pm 9.3$ & $68.8 \pm 9.1$ & $51.6 \pm 9.9$ & $59.2 \pm 9.1$ & $65.9 \pm 10.8$ & $63.1 \pm 9.5$\\
\;\;+ VisPCO & $84.8$ & $69.4$ & $67.6$ & $71.2$ & $77.1$ & $58.1$ & $67.8$ & $75.9$ & $71.5$\\
$\llcorner$ SparseVLM & $ 75.9 \pm 9.8$ & $62.6 \pm 8.2$ & $63.1 \pm 7.2$ & $63.9 \pm 8.6$ & $69.9 \pm 8.2$ & $51.9 \pm 8.3$ & $62.4 \pm 6.9$ & $66.6 \pm 9.8$ & $64.5 \pm 8.4$\\
\;\;+ VisPCO & $85.2$ & $69.0$ & $68.1$ & $71.9$ & $77.6$ & $58.4$ & $67.9$ & $76.3$ & $71.8$\\
$\llcorner$ FitPrune & $ 77.1 \pm 8.7$ & $63.4 \pm 7.7$ & $63.9 \pm 6.8$ & $64.5 \pm 8.2$ & $70.8 \pm 7.9$ & $52.8 \pm 8.1$ & $63.3 \pm 6.4$ & $67.6 \pm 9.4$ & $65.4 \pm 7.9$\\
\;\;+ VisPCO & $85.9$ & $69.4$ & $68.4$ & $72.4$ & $77.9$ & $58.8$ & $68.2$ & $76.6$ & $72.2$\\
\noalign{
  \hrule height 1pt
}
\rowcolor{gray!20}
\multicolumn{10}{c}{\textit{Reduce FLOPs Budget to 10\%, $\sim$0.36 TFLOPs}} \\
$\llcorner$ FastV & $ 33.3 \pm 2.3$ & $30.4 \pm 1.6$ & $44.5 \pm 2.7$ & $33.0 \pm 2.5$ & $39.7 \pm 1.4$ & $29.8 \pm 4.1$ & $8.3 \pm 2.1$ & $33.7 \pm 2.8$ & $31.6 \pm 2.4$\\
\;\;+ VisPCO & $35.5$ & $31.7$ & $46.9$ & $35.5$ & $40.1$ & $33.2$ & $10.1$ & $36.1$ & $33.6$\\
$\llcorner$ SparseVLM & $ 33.6 \pm 2.1$ & $31.2 \pm 1.3$ & $44.9 \pm 2.5$ & $33.9 \pm 2.3$ & $40.3 \pm 1.1$ & $30.5 \pm 3.7$ & $9.1 \pm 2.0$ & $34.4 \pm 2.2$ & $32.2 \pm 2.2 $\\
\;\;+ VisPCO & $35.5$ & $31.5$ & $47.1$ & $35.8$ & $40.4$ & $33.3$ & $10.2$ & $36.3$ & $33.8$\\
$\llcorner$ FitPrune & $ 33.8 \pm 2.1$ & $31.5 \pm 1.1$ & $45.3 \pm 2.4$ & $34.2 \pm 2.2$ & $40.6 \pm 1.0$ & $30.9 \pm 3.5$ & $9.6 \pm 1.9$ & $34.6 \pm 2.1$ & $32.6 \pm 2.0 $\\
\;\;+ VisPCO & $35.6$ & $31.6$ & $47.3$ & $35.8$ & $40.9$ & $33.5$ & $10.4$ & $36.4$ & $33.9$\\
\noalign{
  \hrule height 1pt
}
\end{tabular}}
\vspace{-1em}
\end{table*}

For the second case, we explore several kernel functions to capture diverse pruning patterns. First, inspired by the Ebbinghaus forgetting curve from cognitive psychology~\citep{ebbinghaus}, we design an exponential decay kernel to investigate whether VLMs exhibit similar attention decay patterns for visual tokens across layers:

\begin{equation}
\mathcal{K}_{\text{e}}(i; k, r) = r \cdot e^{-k \cdot i}.
\label{eq:kernel_exp}
\end{equation}

Second, we consider a linear decay kernel that models uniform, gradual token reduction:

\begin{equation}
\mathcal{K}_{\text{l}}(i; k, r) = -k \cdot i + r.
\label{eq:kernel_lin}
\end{equation}

Third, motivated by findings in~\citep{G-Search} showing that attention score rankings remain similar across layers and follow a sigmoid-like curve, we adopt a gentle p-sigmoid kernel. Following Eq.~\eqref{eq:kernel_single}, we use a smaller $\gamma$ to capture smooth, progressive pruning transitions. Finally, inspired by hierarchical representation learning where deep networks perform feature extraction at different levels, we introduce a multi-step sigmoid kernel to model the hypothesis that VLMs compress information at multiple critical layers:

\begin{footnotesize}
\begin{equation}
\mathcal{K}_{\text{ms}}(i; k, r, M) = 1 - \sum_{j=1}^{M} \frac{1-r}{M} \cdot \sigma\left(k\left(i - \frac{(2j-1)L}{2M}\right)\right),
\label{eq:kernel_ms}
\end{equation}    
\end{footnotesize}
where $M$ is the number of pruning steps, and the $j$-th step is centered at layer $\frac{(2j-1)L}{2M}$, which evenly distributes the steps across layers. This design creates $M$ evenly-spaced decision points for progressive information compression across layers. Collectively, these learnable kernels capture a wide spectrum of pruning patterns, enabling the model to discover task-specific compression strategies. 

The parameters $k$ and $r$ are dynamically predicted by a lightweight surrogate neural network $f_{\theta}$. As illustrated in the top-right of Figure~\ref{fig:architecture}, the network takes as input the concatenated visual and textual embeddings along with the computational budget $B$, and computes layer-wise retention ratios $r_i = \mathcal{K}(i; k, r)$ according to the selected pruning pattern. All outputs $r_i$ are clipped to $[0, 1]$. Through gradient-based optimization, the surrogate network learns which layers and tokens are critical. This provides mechanistic insights into how VLMs prioritize visual information across layers.

\section{Experiments}
\label{sec: experiments}
\subsection{Implementation Details}
\label{sec: experiments_part_0}
We use Qwen2.5VL-3B~\citep{Qwen25VL} as the base model and construct our training set by downsampling 30K samples from LLaVA-Instruct-150K~\citep{LLaVA1.5}. To mitigate the long-tail distribution of image resolutions, we apply resolution-based resampling to prevent performance degradation on rare image sizes. Our evaluation spans three categories of benchmarks: visual question answering (A-OKVQA~\citep{AOKVQA}, VizWiz~\citep{VizWiz}, SEEDBench~\citep{SeedBench}), multimodal reasoning (MMBench~\citep{MMBench}, MME~\citep{MME}), and chart understanding (ChartQA~\citep{ChartQA}, OCRBench~\citep{OCRBench}, TextVQA~\citep{TextVQA}).

For single-layer pruning, we set the penalty parameter $\lambda=100$, convergence threshold $\epsilon=0.01$, and update coefficients $\alpha=2$, $\beta=0.5$. The Gaussian kernel width $\sigma=10$ and temperature $T=0.1$ control the continuous relaxation and straight-through estimator, respectively. We use the AdamW optimizer with a learning rate of $4 \times 10^{-4}$ and batch size of 16. All experiments are conducted on 8 NVIDIA H20 GPUs (96GB each). Other training configurations are provided in Appendix~\ref{Appendix: Experiment Settings}.

\subsection{Pareto Frontier Approximation}
\label{sec: experiments_part_1}
\begin{figure*}[!t]
    \centering
    \begin{subfigure}[b]{0.32\textwidth}
        \centering
        \includegraphics[width=\textwidth]{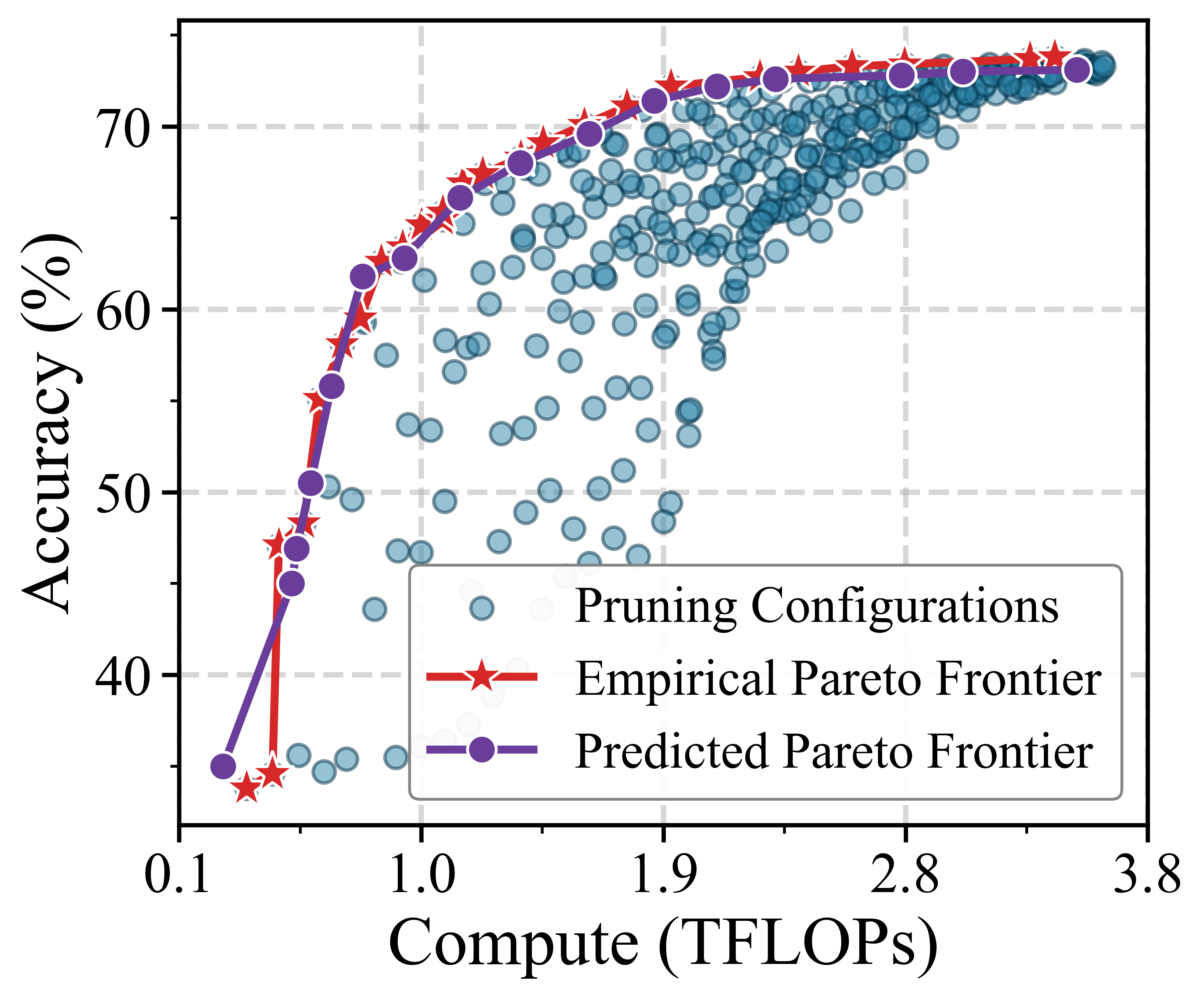}
        \label{fig:exp1_fig}
    \end{subfigure}
    \hfill
    \begin{subfigure}[b]{0.32\textwidth}
        \centering
        \includegraphics[width=\textwidth]{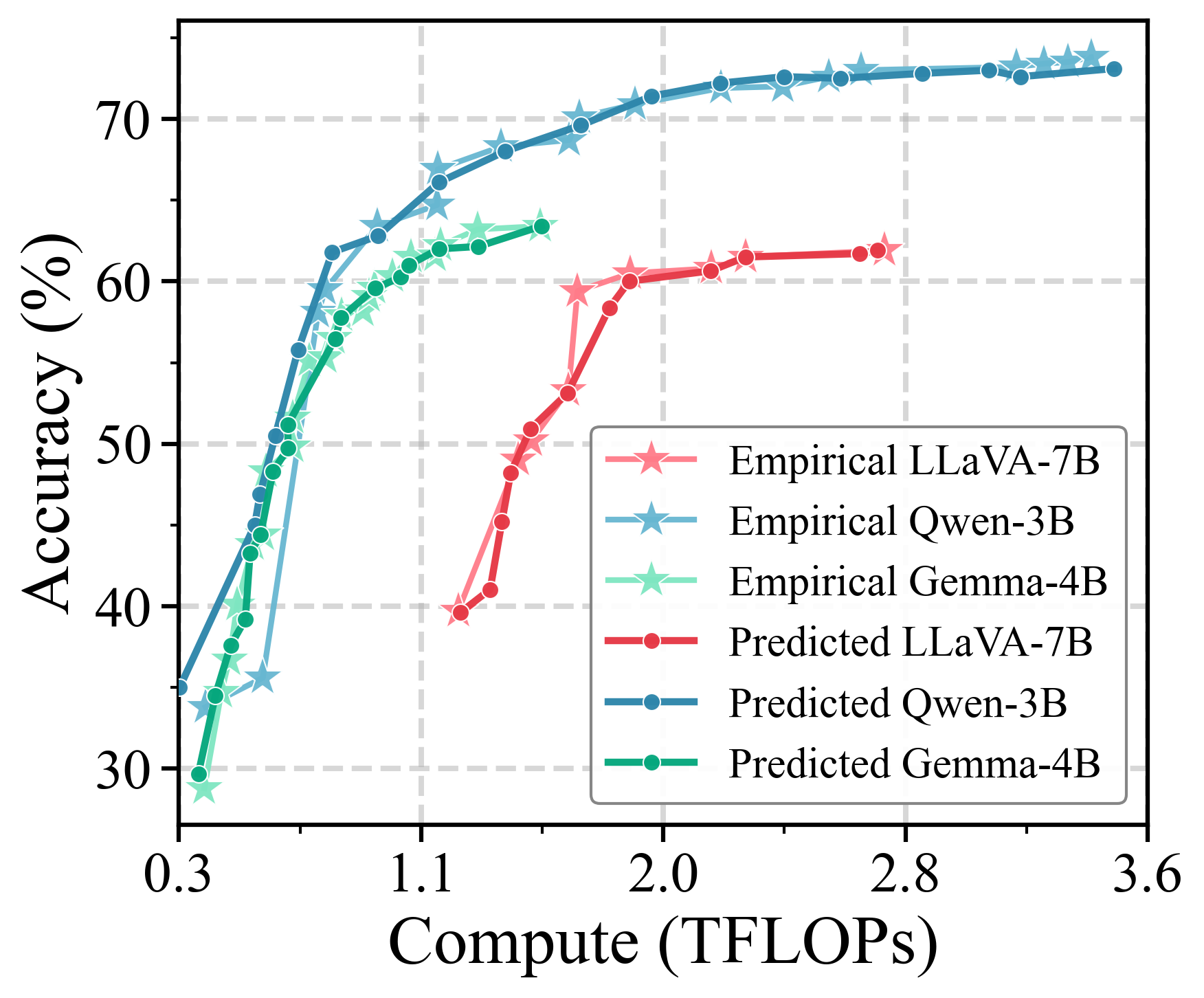}
        \label{fig:exp2_fig}
    \end{subfigure}
    \hfill
    \begin{subfigure}[b]{0.32\textwidth}
        \centering
        \includegraphics[width=\textwidth]{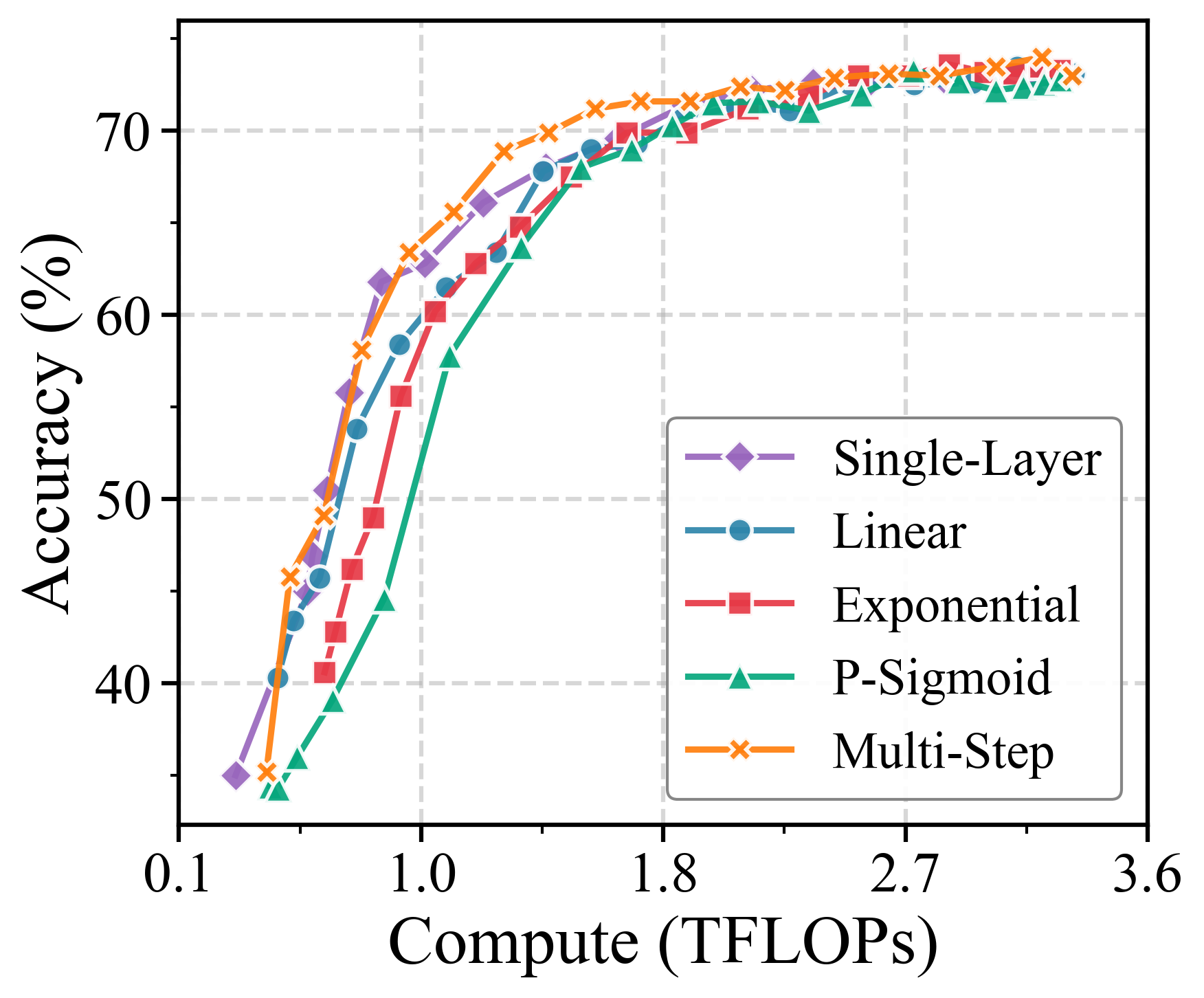}
        \label{fig:exp3_fig}
    \end{subfigure}
    \vspace{-1em}
    \caption{Experimental results of \textbf{VisPCO}. (Left) Comparison between empirical and predicted Pareto frontiers. (Middle) Comparison between empirical and predicted Pareto frontiers across different VLM architectures. (Right) Comparison of Pareto frontiers among different pruning patterns.}
    \label{fig:all_exp}
    \vspace{-1em}
\end{figure*}

We apply \textbf{VisPCO} to three representative pruning methods: FastV~\citep{FastV}, FitPrune~\citep{FitPrune}, and SparseVLM~\citep{SparseVLM}, which employ different importance scoring mechanisms for visual tokens. Table~\ref{tab: exp1} compares performance before and after applying \textbf{VisPCO} under different FLOPs budgets. For each method, results without \textbf{VisPCO} are obtained by sampling multiple configurations and averaging their performance, with standard deviations reported ($\pm$ std). Figure~\ref{fig:all_exp}(left) illustrates the empirical Pareto frontier obtained through grid search and the predicted Pareto frontier by \textbf{VisPCO}, with computational budget on the x-axis and average accuracy across 8 visual benchmarks on the y-axis.

\textbf{Performance gains at moderate budgets.} As shown in Table~\ref{tab: exp1}, \textbf{VisPCO}'s benefits vary significantly across different computational budget regimes. At extreme budgets, configuration selection has limited impact. For example, at 90\% budget, performance varies by less than 1 percentage point across different configurations, as resources are abundant enough that most configurations perform well. Similarly, at very low budgets, severe resource constraints limit all configurations. In contrast, moderate budgets (e.g., 50\%) present a critical regime where configuration choice significantly impacts performance---different configurations can vary by up to 19 percentage points. This substantial performance gap demonstrates the importance of principled configuration optimization and validates the need for methods like \textbf{VisPCO}.

\textbf{Quality of frontier approximation.} As shown in Figure~\ref{fig:all_exp}(left), the predicted Pareto frontiers exhibit near-perfect alignment with empirical frontiers. This validates the effectiveness of our kernel-based approximation approach in capturing the true computation-performance trade-off landscape. Table~\ref{tab: exp1_2} presents a comprehensive comparison with existing approaches, including predefined pruning configuration strategies, training-based methods, and random search baselines (Random-$\mathrm{N}$ denotes selecting the best from $\mathrm{N}$ random samples). We evaluate both search efficiency and performance. \textbf{VisPCO} outperforms all baseline methods including VTW~\citep{VTW}, G-Search~\citep{G-Search}, ATP-LLaVA~\citep{APT-LLaVA}, MADTP~\citep{MADTP}, AIM~\citep{AIM}, and moderate random search variants, while requiring only 1 hour of training time. Notably, MADTP requires additional training of MAG and DTP modules (6+ hours) and incurs substantially higher FLOPs (3.91T). AIM is training-free with a fixed layer-wise pruning strategy; at comparable FLOPs (2.20T vs.\ 2.33T), \textbf{VisPCO} achieves clearly superior performance (MMB: 43.6 vs.\ 39.5; TQA: 81.3 vs.\ 74.6; SEED: 47.5 vs.\ 43.8).

\let\thefootnote\relax\footnotetext{$^{\dagger}$MME scores are normalized to percentages.}

\begin{table}[!t]
\centering
\caption{Comparison of configuration search methods. Time represents search time to identify optimal configurations (training time for \textbf{VisPCO}). Random-$\mathrm{N}$ denotes random sampling with $\mathrm{N}$ evaluations. All methods target similar computational budgets.}
\vspace{-0.5em}
\label{tab: exp1_2}
\resizebox{\columnwidth}{!}{
\begin{tabular}{l|c|c|ccc}
\toprule[0.1em]
\multicolumn{1}{c|}{\textbf{Methods}} & \multicolumn{1}{c|}{\textbf{FLOPs (T) $\downarrow$}} & \textbf{Time (h) $\downarrow$} & \textbf{MMB $\uparrow$} & \textbf{SEED $\uparrow$} & \multicolumn{1}{c}{\textbf{TQA $\uparrow$}}\\
\midrule[0.1em]
VTW & $ 2.34 $ & ${1}$+ & $ 36.5 $ & $ 44.1 $ & $ 73.2 $\\
G-Search & $ 2.58 $ & - & $ 42.1 $ & $ 47.5 $ & $ 80.2 $\\
ATP-LLaVA & $ 2.23 $ & $48$+ & $ 37.2 $ & $ 45.5 $ & $ 77.2 $\\
MADTP & $ 3.91 $ & $6$+ & $ 42.4 $ & $ 46.8 $ & $ 79.1 $\\
AIM & $ 2.33 $ & - & $ 39.5 $ & $ 43.8 $ & $ 74.6 $\\
\midrule
Random-40 & $2.18$ & $12$+ & $31.2$ & $36.6$ & $70.9$\\
Random-80 & $2.20$ & $24$+ & $42.8$ & $47.7$ & $80.1$\\
Random-160 & $2.20$ & $48$+ & $44.0$ & $48.1$ & $82.4$\\
\midrule
\textbf{VisPCO} & ${2.20}$ & ${1}$+ & ${43.6}$ & ${47.5}$ & ${81.3}$\\
\bottomrule[0.1em]
\end{tabular}
}
\vspace{-1.5em}
\end{table}

\subsection{Cross-Model Generalization}
\label{sec: experiments_part_2}
To validate the generalization capability of \textbf{VisPCO} across different VLM architectures, we apply it to Gemma3-4B~\citep{Gemma3} and LLaVA-v1.5-7B~\citep{LLaVA1.5}. Table~\ref{tab: exp2} presents representative results using FastV~\citep{FastV} under a 50\% FLOPs budget.

\begin{table}[!t]
\centering
\caption{Performance of different VLMs with and without \textbf{VisPCO} at 50\% FLOPs budget ($\pm$ std).}
\vspace{-0.5em}
\label{tab: exp2}
\resizebox{\columnwidth}{!}{
\begin{tabular}{l|ccc|c}
\toprule[0.1em]
\multicolumn{1}{l|}{\textbf{Models}} & \textbf{AOKVQA} & \textbf{MMBench} & \multicolumn{1}{c|}{\textbf{TextVQA}} & \textbf{Avg (\%)}\\
\midrule[0.1em]
LLaVA-7B & $50.3 \pm 10.7$ & $26.5 \pm 11.5$ & $64.2 \pm 11.4$ & $47.0 \pm 11.2$ \\
\;+ VisPCO & $60.8$ & $38.0$ & $75.2$ & $58.0$ \\
\midrule
Gemma3-4B & $41.2 \pm 12.4$ & $44.7 \pm 12.9$ & $33.6 \pm 12.3$ & $39.8 \pm 12.5$ \\
\;+ VisPCO & $53.4$ & $57.3$ & $45.5$ & $52.1$ \\
\midrule
Qwen2.5VL-3B & $74.7 \pm 10.1$ & $62.4 \pm 9.3$ & $65.9 \pm 10.8$ & $67.7 \pm 10.1$ \\
\;+ VisPCO & $84.8$ & $71.2$ & $75.9$ & $77.3$ \\
\bottomrule[0.1em]
\end{tabular}
}
\end{table}

\begin{table}[!t]
\centering
\caption{Hardware performance measurements at 50\% compute budget across models. TTFT: time to first token. Throughput: tokens generated per second. Avg.\ Perf.: average accuracy across benchmarks.}
\vspace{-0.5em}
\label{tab: hardware}
\resizebox{\columnwidth}{!}{
\begin{tabular}{l|c|c|c|c}
\toprule[0.1em]
\textbf{Method} & \textbf{Budget} & \textbf{TTFT (ms) $\downarrow$} & \textbf{Throughput (tokens/s) $\uparrow$} & \textbf{Avg.\ Perf.\ $\uparrow$} \\
\midrule[0.1em]
Qwen2.5VL-3B & 100\% & $83 \pm 3$ & $18 \pm 2$ & $78.1$ \\
\;\;+ FastV & 50\% & $74 \pm 2$ & $20 \pm 3$ & $63.1$ \\
\;\;+ FastV + VisPCO & 50\% & $76 \pm 3$ & $20 \pm 2$ & $\mathbf{71.5}$ \\
\midrule
LLaVA-v1.5-7B & 100\% & $114 \pm 12$ & $14 \pm 6$ & $63.9$ \\
\;\;+ FastV & 50\% & $96 \pm 10$ & $16 \pm 6$ & $41.4$ \\
\;\;+ FastV + VisPCO & 50\% & $97 \pm 8$ & $16 \pm 4$ & $\mathbf{52.3}$ \\
\midrule
Gemma3-4B & 100\% & $94 \pm 6$ & $16 \pm 4$ & $68.9$ \\
\;\;+ FastV & 50\% & $81 \pm 7$ & $18 \pm 5$ & $38.8$ \\
\;\;+ FastV + VisPCO & 50\% & $81 \pm 6$ & $18 \pm 5$ & $\mathbf{50.8}$ \\
\bottomrule[0.1em]
\end{tabular}
}
\vspace{-1.5em}
\end{table}

\textbf{Consistency across architectures.} While Qwen2.5VL exhibits a wider performance range compared to Gemma3 and LLaVA, \textbf{VisPCO} consistently selects optimal configurations. Figure~\ref{fig:all_exp}(middle) shows that predicted Pareto frontiers closely match the empirical frontiers across all architectures, demonstrating robust generalization. Notably, Qwen2.5VL's frontier is positioned in the upper-left region, indicating superior accuracy-efficiency characteristics. This advantage stems from its native image resolution, whereas Gemma3 and LLaVA resize inputs to fixed dimensions. This suggests preserving original dimensions with learned pruning is potentially more effective than aggressive preprocessing for efficient VLMs.

\textbf{Hardware efficiency.} To verify that FLOPs reductions translate into practical speedups, we measure time-to-first-token (TTFT) and throughput on an NVIDIA H20 GPU. As shown in Table~\ref{tab: hardware}, at a 50\% compute budget, FastV with \textbf{VisPCO} maintains the same hardware efficiency as FastV alone---comparable TTFT and throughput---while recovering substantial performance lost from pruning. For example, on Qwen2.5VL-3B, \textbf{VisPCO} improves average performance from 63.1\% to 71.5\% with negligible latency overhead (76 vs.\ 74 ms TTFT). Similar patterns hold across LLaVA-v1.5-7B and Gemma3-4B, confirming that \textbf{VisPCO} preserves the underlying pruning method's hardware efficiency while substantially improving task performance.

\subsection{Analysis of Pruning Patterns}
\label{sec: experiments_part_3}
We investigate the performance differences between single-layer and multi-layer pruning strategies, as well as the impact of different kernel choices in multi-layer configurations. Table~\ref{tab: exp3} presents the Pareto-optimal results for different pruning patterns under a 50\% computational budget. Figure~\ref{fig:all_exp}(right) illustrates and compares the Pareto frontiers across different pruning patterns.

\begin{table}[!t]
\centering
\caption{Comparison of performance under different pruning patterns at 50\% FLOPs budget.}
\vspace{-0.5em}
\label{tab: exp3}
\small
\setlength{\tabcolsep}{4pt}
\resizebox{0.5\textwidth}{!}{
\begin{tabular}{l|c|ccc|c}
\toprule[0.1em]
\multicolumn{1}{l|}{\textbf{Patterns}} & \textbf{Kernels} & \textbf{AOK} & \textbf{MMB} & \multicolumn{1}{c|}{\textbf{TQA}} & \textbf{Avg (\%)}\\
\midrule[0.1em]
\multicolumn{1}{l|}{Single-Layer} & - & $84.8$ & $71.2$ & $75.9$ & $77.3$ \\
\midrule
& Linear & $82.6$ & $70.9$ & $74.9$ & $76.1$ \\
\multirow{4}{*}[1em]{Multi-Layer} & Exponential & $82.2$ & $70.4$ & $74.4$ & $75.7$ \\
& P-Sigmoid & $81.9$ & $69.6$ & $74.1$ & $75.2$ \\
& Multi-Step & $84.9$ & $71.8$ & $76.7$ & $77.8$ \\
\bottomrule[0.1em]
\end{tabular}
}
\vspace{-1em}
\end{table}

\textbf{Strategic pruning pattern selection.} As shown in Table~\ref{tab: exp3}, multi-layer pruning with the multi-step kernel achieves the best performance under a 50\% budget, outperforming both single-layer pruning and other kernel variants (linear, exponential, sigmoid). Figure~\ref{fig:all_exp}(right) reveals that this advantage is budget-dependent. When computational budget exceeds 50\%, all pruning patterns converge to comparable performance and closely approximate the empirical Pareto frontier, making strategy selection less critical. However, below 50\% budget, notable differences emerge among pruning patterns, with the multi-step kernel showing clear superiority.

\textbf{Implications for VLM design.} The multi-step kernel's superior performance at low budgets reveals important architectural insights. Visual token redundancy emerges at specific layers rather than uniformly across the network. Certain layers introduce redundancy through attention or feature transformations, while others preserve essential representations. The multi-step kernel identifies these critical compression points, enabling targeted pruning while retaining key information. These findings provide practical guidance. When resources are sufficient (budget >50\%), simple single-layer pruning achieves near-optimal performance. Under tight constraints (budget <50\%), multi-step layer-wise pruning is recommended to better exploit VLMs' hierarchical compression structure.

\section{Conclusion}
\label{sec: conclusion}
In this paper, we introduced \textbf{VisPCO}, a novel computation budget-aware framework for automatically optimizing visual token pruning configurations in vision-language models. By formulating the problem as Pareto optimization with continuous relaxation, \textbf{VisPCO} enables efficient end-to-end gradient-based training to automatically identify optimal pruning configurations for any given computational budget. This approach significantly reduces search costs compared to traditional exhaustive grid search methods. Extensive experiments across 8 visual benchmarks demonstrate that our method generalizes well across various pruning strategies and VLM architectures. Furthermore, our investigation through learnable kernel functions reveals that progressive multi-step pruning consistently outperforms both single-layer and other multi-layer kernel approaches, providing valuable insights for efficient VLM design in resource-constrained deployment scenarios.

\section*{Limitations}
Although our framework demonstrates strong performance across diverse benchmarks and model architectures, several limitations remain to be addressed in future work. First, our experiments primarily focus on single-image tasks; further validation is needed to assess how effectively our optimized pruning configurations generalize to multi-image and video inputs, which involve more complex temporal and spatial redundancies. Second, our proposed kernel functions provide a structured and interpretable way to model pruning distributions. Future work could explore extending this approach to learn more flexible, non-parametric or input-adaptive patterns, potentially capturing even more nuanced task-specific pruning strategies.

\section*{Ethics Statement}
This work focuses on optimizing visual token pruning configurations for vision-language models to improve computational efficiency. Our method does not involve the collection or use of private or sensitive data; all experiments are conducted on publicly available benchmarks. We do not foresee direct negative societal impacts from this research. By reducing the computational cost of VLMs, our work may contribute to lowering energy consumption and carbon emissions associated with large-scale model inference, thereby promoting more sustainable and accessible AI deployment.

\section*{Acknowledgments}
We sincerely thank the students and engineers at the Data Intelligence Research Center, Shanghai Jiao Tong University, for their assistance during the development of this work. This work was supported by NSF China under Grant No.T2421002,  92579104, 62525209, T2542021.

\bibliography{custom}

\newpage
\appendix

\section{FLOPs Computation}
\label{Appendix: FLOPs Analysis}
In this section, we provide a detailed derivation of the floating-point operations (FLOPs) computation for Transformer layers in vision-language models.

For a standard Transformer layer, the primary computational costs come from the self-attention mechanism and the feed-forward network (FFN). Given a sequence length $N$ and hidden dimension $D$, we compute the FLOPs for each component separately.

\subsection{Self-Attention Mechanism}
The self-attention mechanism consists of the following operations:

\textbf{(1) Linear projections:} Three projection matrices $\mathbf{W}_Q, \mathbf{W}_K, \mathbf{W}_V \in \mathbb{R}^{D \times D}$ map the input to Query, Key, and Value representations. Each matrix multiplication requires $2ND^2$ floating-point operations (multiplying input of size $N \times D$ with weight of size $D \times D$), thus:
\begin{equation}
    \text{FLOPs}_{\text{QKV}} = 3 \times 2ND^2 = 6ND^2.
\end{equation}

\textbf{(2) Attention score:} Computing $\mathbf{Q}\mathbf{K}^T \in \mathbb{R}^{N \times N}$, where $\mathbf{Q}, \mathbf{K} \in \mathbb{R}^{N \times D}$:
\begin{equation}
    \text{FLOPs}_{\text{Score}} = 2N^2D.
\end{equation}

\textbf{(3) Attention weighting:} Computing $\text{Softmax}(\mathbf{Q}\mathbf{K}^T / \sqrt{D})\mathbf{V}$, i.e., multiplying $\mathbb{R}^{N \times N}$ with $\mathbb{R}^{N \times D}$:
\begin{equation}
    \text{FLOPs}_{\text{Weight}} = 2N^2D.
\end{equation}

Note: The FLOPs for the Softmax operation are relatively small and typically neglected.

\textbf{(4) Output projection:} Projecting back to the original dimension through $\mathbf{W}_O \in \mathbb{R}^{D \times D}$:
\begin{equation}
    \text{FLOPs}_{\text{Output}} = 2ND^2.
\end{equation}

Therefore, the total FLOPs for the self-attention mechanism is:
\begin{equation}
    \text{FLOPs}_{\text{Attention}} = 8ND^2 + 4N^2D.
\end{equation}

\subsubsection{Feed-Forward Network}
The standard FFN consists of two linear layers with an intermediate dimension $D_{\text{ffn}}$:
\begin{align}
    \text{FFN}(\mathbf{x}) = \mathbf{W}_2 \cdot \text{GELU}(\mathbf{W}_1 \cdot \mathbf{x}).
\end{align}

where $\mathbf{W}_1 \in \mathbb{R}^{D \times D_{\text{ffn}}}$ and $\mathbf{W}_2 \in \mathbb{R}^{D_{\text{ffn}} \times D}$.
The total FLOPs for the FFN is:
\begin{equation}
    \text{FLOPs}_{\text{FFN}} = 4ND_{\text{ffn}}D.
\end{equation}

\subsubsection{Total FLOPs per Layer}
Combining self-attention and FFN, the total FLOPs for a single Transformer layer is:
\begin{equation}
    \text{FLOPs}_{\text{layer}} = 8ND^2 + 4N^2D + 4ND_{\text{ffn}}D.
\end{equation}

In standard Transformer architectures, $D_{\text{ffn}} = 4D$, which gives:
\begin{equation}
    \text{FLOPs}_{\text{layer}} = 24ND^2 + 4N^2D.
\end{equation}
We neglect relatively small computational costs such as LayerNorm and residual connections.

\subsection{Total FLOPs for Vision-Language Models}

For vision-language models, the input sequence consists of text tokens and visual tokens. Let $N_t$ denote the number of text tokens and $N_v$ denote the initial number of visual tokens. The total number of tokens at layer $i$ is:
\begin{equation}
    N_i = N_t + r_iN_v
\end{equation}
where $r_i \in [0,1]$ represents the retention ratio of visual tokens at layer $i$.

For a Transformer model with $L$ layers, the total computational cost is:

\begin{small}
\begin{equation}
F(\mathbf{r}) = \sum_{i=1}^{L}\left[24(N_t + r_iN_v)D^2 + 4(N_t + r_iN_v)^2D\right]
\end{equation}
\end{small}

where:
\begin{itemize}
    \item The first term $24(N_t + r_iN_v)D^2$ corresponds to linear projections in self-attention and the FFN
    \item The second term $4(N_t + r_iN_v)^2D$ corresponds to the quadratic complexity of attention matrix computation
    \item $\mathbf{r} = [r_1, r_2, \ldots, r_L]$ is the vector of visual token retention ratios across layers
\end{itemize}

This formula indicates that as visual tokens are pruned ($r_i$ decreases), the model's computational cost is significantly reduced, especially the quadratic complexity term.

\section{Theoretical Analysis}
\subsection{Proof of Theorem~\ref{thm:convergence}}
\label{Appendix:proof_theorem1}

Consider the equality-constrained optimization problem:
\begin{equation}
\begin{aligned}
\min \quad & f(\mathbf{x}) \\
\text{s.t.} \quad & h_j(\mathbf{x}) = 0, \quad j = 1,\ldots,l,
\end{aligned}
\label{eq:original_problem}
\end{equation}
where $f, h_j: \mathbb{R}^n \rightarrow \mathbb{R}$ are twice continuously differentiable functions. The augmented Lagrangian function for this problem is given by Equation~\eqref{eq:augmented_lagrangian_function}.

Let $\bar{\mathbf{x}}$ be a local optimal solution of problem~\eqref{eq:original_problem} that satisfies the second-order sufficient conditions. That is, there exists a Lagrange multiplier vector $\bar{\mathbf{v}} = [\bar{v}_1, \ldots, \bar{v}_l]^T$ such that:
\begin{equation}
\nabla f(\bar{\mathbf{x}}) - \mathbf{A}\bar{\mathbf{v}} = 0,
\label{eq:first_order}
\end{equation}
\begin{equation}
h_j(\bar{\mathbf{x}}) = 0, \quad j = 1,\ldots,l,
\label{eq:feasibility}
\end{equation}
and for every nonzero vector $\mathbf{d}$ satisfying $\mathbf{d}^T \nabla h_j(\bar{\mathbf{x}}) = 0$ for $j=1,\ldots,l$, we have:
\begin{equation}
\mathbf{d}^T \nabla^2_{\mathbf{x}} \mathcal{L}(\bar{\mathbf{x}}, \bar{\mathbf{v}}) \mathbf{d} > 0,
\label{eq:second_order}
\end{equation}
where
\begin{equation}
\mathbf{A} = [\nabla h_1(\bar{\mathbf{x}}), \ldots, \nabla h_l(\bar{\mathbf{x}})],
\label{eq:jacobian}
\end{equation}
and $\mathcal{L}(\mathbf{x}, \mathbf{v}) = f(\mathbf{x}) - \mathbf{v}^T \mathbf{h}(\mathbf{x})$ is the standard Lagrangian function.

By assumption, $\bar{\mathbf{x}}$ is a Karush-Kuhn-Tucker (KKT) point of problem~\eqref{eq:original_problem}, thus:
\begin{equation}
\nabla_{\mathbf{x}} \phi(\bar{\mathbf{x}}, \bar{\mathbf{v}}, \lambda) = 0.
\label{eq:gradient_zero}
\end{equation}

We now prove that the Hessian matrix $\nabla^2_{\mathbf{x}} \phi(\bar{\mathbf{x}}, \bar{\mathbf{v}}, \lambda)$ is positive definite at $\bar{\mathbf{x}}$ for sufficiently large $\lambda$.

From Equation~\eqref{eq:augmented_lagrangian_function}, we can derive:

\begin{small}
\begin{align}
&\nabla^2_{\mathbf{x}} \phi(\mathbf{x}, \bar{\mathbf{v}}, \lambda) 
= \nabla^2 f(\mathbf{x}) - \sum_{j=1}^l \bar{v}_j \nabla^2 h_j(\mathbf{x}) \nonumber \\
&+ \sigma \sum_{j=1}^l h_j(\mathbf{x}) \nabla^2 h_j(\mathbf{x}) + \lambda \sum_{j=1}^l \nabla h_j(\mathbf{x}) \nabla h_j(\mathbf{x})^T \\
&= \nabla^2 f(\mathbf{x}) - \sum_{j=1}^l (\bar{v}_j - \lambda h_j(\mathbf{x})) \nabla^2 h_j(\mathbf{x}) \nonumber \\
&+ \lambda \sum_{j=1}^l \nabla h_j(\mathbf{x}) \nabla h_j(\mathbf{x})^T = \mathbf{Q} + \lambda \mathbf{A}\mathbf{A}^T \nonumber,
\label{eq:hessian_decomposition}
\end{align}  
\end{small}
where
\begin{equation}
\mathbf{Q} = \nabla^2 f(\mathbf{x}) - \sum_{j=1}^l (\bar{v}_j - \lambda h_j(\mathbf{x})) \nabla^2 h_j(\mathbf{x}),
\end{equation}
\begin{equation}
\mathbf{A} = [\nabla h_1(\mathbf{x}), \ldots, \nabla h_l(\mathbf{x})].
\end{equation}

At the point $\bar{\mathbf{x}}$, we have:
\begin{equation}
\nabla^2_{\mathbf{x}} \phi(\bar{\mathbf{x}}, \bar{\mathbf{v}}, \lambda) = \bar{\mathbf{Q}} + \lambda \bar{\mathbf{A}}\bar{\mathbf{A}}^T,
\label{eq:hessian_at_xbar}
\end{equation}
where $\bar{\mathbf{Q}}$ and $\bar{\mathbf{A}}$ denote the evaluations at $\bar{\mathbf{x}}$.

Let $\text{rank}(\bar{\mathbf{A}}) = r \leq l$, and let $\mathbf{B} \in \mathbb{R}^{n \times r}$ be an orthonormal basis matrix for $\bar{\mathbf{A}}$ (i.e., $\mathbf{B}^T\mathbf{B} = \mathbf{I}_r$), meaning the $r$ columns of $\mathbf{B}$ form an orthonormal basis for the subspace spanned by the $l$ columns of $\bar{\mathbf{A}}$. Thus, we have:
\begin{equation}
\bar{\mathbf{A}} = \mathbf{B}\mathbf{C},
\label{eq:basis_decomposition}
\end{equation}
where $\mathbf{C} = \mathbf{B}^T\bar{\mathbf{A}}$ has rank $r$.

For any nonzero vector $\mathbf{u} \in \mathbb{R}^n$, we decompose it as:
\begin{equation}
\mathbf{u} = \mathbf{p} + \mathbf{B}\mathbf{q},
\label{eq:vector_decomposition}
\end{equation}
where $\mathbf{p}$ satisfies $\mathbf{B}^T\mathbf{p} = \mathbf{0}$. Clearly, $\bar{\mathbf{A}}^T\mathbf{p} = \mathbf{0}$, which implies:
\begin{equation}
\nabla h_j(\bar{\mathbf{x}})^T \mathbf{p} = 0, \quad j = 1,\ldots,l.
\label{eq:orthogonality}
\end{equation}

Now, we can write $\mathbf{u}^T \nabla^2_{\mathbf{x}} \phi(\bar{\mathbf{x}}, \bar{\mathbf{v}}, \lambda) \mathbf{u}$ as:
\begin{small}
\begin{align}
&\mathbf{u}^T \nabla^2_{\mathbf{x}} \phi(\bar{\mathbf{x}}, \bar{\mathbf{v}}, \lambda) \mathbf{u}\nonumber \\
&= (\mathbf{p} + \mathbf{B}\mathbf{q})^T(\bar{\mathbf{Q}} + \lambda \bar{\mathbf{A}}\bar{\mathbf{A}}^T)(\mathbf{p} + \mathbf{B}\mathbf{q}) \\
&= \mathbf{p}^T\bar{\mathbf{Q}}\mathbf{p} + 2\mathbf{p}^T\bar{\mathbf{Q}}\mathbf{B}\mathbf{q} + \mathbf{q}^T\mathbf{B}^T\bar{\mathbf{Q}}\mathbf{B}\mathbf{q} \nonumber\\
&+ \lambda \mathbf{q}^T\mathbf{C}\mathbf{C}^T\mathbf{q}\nonumber.
\label{eq:quadratic_form}
\end{align}
\end{small}

Since $\bar{\mathbf{x}}$ is a local optimal solution of problem~\eqref{eq:original_problem} satisfying the second-order sufficient conditions, there exists a constant $\alpha > 0$ such that:
\begin{equation}
\mathbf{p}^T\bar{\mathbf{Q}}\mathbf{p} \geq \alpha \|\mathbf{p}\|^2.
\label{eq:positive_definite_p}
\end{equation}

Let $b$ be the largest singular value of $\bar{\mathbf{Q}}\mathbf{B}$, let $e = \|\mathbf{B}^T\bar{\mathbf{Q}}\mathbf{B}\|_2$, and let $\mu > 0$ be the smallest eigenvalue of $\mathbf{C}\mathbf{C}^T$. Then:
\begin{small}
\begin{equation}
\mathbf{u}^T \nabla^2_{\mathbf{x}} \phi(\bar{\mathbf{x}}, \bar{\mathbf{v}}, \lambda) \mathbf{u} \geq \alpha \|\mathbf{p}\|^2 - 2b\|\mathbf{p}\|\|\mathbf{q}\| + (\lambda\mu - e)\|\mathbf{q}\|^2.
\label{eq:lower_bound}
\end{equation} 
\end{small}

Since $\mathbf{u} \neq \mathbf{0}$, the vectors $\mathbf{p}$ and $\mathbf{q}$ cannot both be zero. Therefore, if we choose $\lambda$ sufficiently large such that:
\begin{equation}
\lambda\mu - e - \frac{b^2}{\alpha} > 0,
\label{eq:sigma_condition}
\end{equation}
that is,
\begin{equation}
\lambda > \frac{b^2 + \alpha e}{\alpha \mu},
\label{eq:sigma_threshold}
\end{equation}
then we always have:
\begin{equation}
\mathbf{u}^T \nabla^2_{\mathbf{x}} \phi(\bar{\mathbf{x}}, \bar{\mathbf{v}}, \lambda) \mathbf{u} > 0.
\label{eq:positive_definite}
\end{equation}

Therefore, there exists:
\begin{equation}
\lambda' = \frac{b^2 + \alpha e}{\alpha \mu}.
\label{eq:sigma_prime}
\end{equation}

When the penalty parameter $\lambda > \lambda'$, the matrix $\nabla^2_{\mathbf{x}} \phi(\bar{\mathbf{x}}, \bar{\mathbf{v}}, \lambda)$ is positive definite. Combined with Equations~\eqref{eq:gradient_zero} and~\eqref{eq:positive_definite}, we conclude that $\bar{\mathbf{x}}$ is a strict local minimizer of $\phi(\mathbf{x}, \bar{\mathbf{v}}, \lambda)$. This completes the proof.

\subsection{Elimination of $y$ via Quadratic Completion}
\label{Appendix: eliminate y}

Starting from the augmented Lagrangian function in Equation~\eqref{eq:our_augmented_lagrangian}, we apply the technique of completing the square to eliminate the auxiliary variable $y$.

Let $g(\mathbf{r}) = B - F(\mathbf{r})$ denote the constraint function. We can rewrite Equation~\eqref{eq:our_augmented_lagrangian} as:
\begin{small}
\begin{align}
&\tilde{\phi}(\mathbf{r}, y, w, \lambda) \nonumber\\
&= \mathcal{L}_{\text{distill}}(\mathbf{r}) - w(g(\mathbf{r}) - y^2) + \frac{\lambda}{2}(g(\mathbf{r}) - y^2)^2  \\
&= \mathcal{L}_{\text{distill}}(\mathbf{r}) + \left[-w(g(\mathbf{r}) - y^2) + \frac{\lambda}{2}(g(\mathbf{r}) - y^2)^2\right]\nonumber.
\end{align}
\end{small}

Completing the square with respect to $y^2$, we have:
\begin{align}
&-w(g(\mathbf{r}) - y^2) + \frac{\lambda}{2}(g(\mathbf{r}) - y^2)^2 \nonumber \\
&= \frac{\lambda}{2}\left[(g(\mathbf{r}) - y^2) - \frac{w}{\lambda}\right]^2 - \frac{w^2}{2\lambda} \nonumber\\
&= \frac{\lambda}{2}\left[y^2 - \left(g(\mathbf{r}) - \frac{w}{\lambda}\right)\right]^2 - \frac{w^2}{2\lambda}.
\label{eq:complete_square}
\end{align}

To minimize $\tilde{\phi}$ with respect to $y$, we analyze the optimal value of $y^2$. The term $\frac{\lambda}{2}\left[y^2 - \left(g(\mathbf{r}) - \frac{w}{\lambda}\right)\right]^2$ is minimized when:
\begin{equation}
y^2 = g(\mathbf{r}) - \frac{w}{\lambda} = \frac{1}{\lambda}(\lambda g(\mathbf{r}) - w).
\label{eq:y_squared_unconstrained}
\end{equation}

However, since $y \in \mathbb{R}$, we must have $y^2 \geq 0$. Therefore, the optimal value is:
\begin{equation}
y^2 = \max\left\{0, \frac{1}{\lambda}(\lambda g(\mathbf{r}) - w)\right\}.
\label{eq:y_squared_optimal}
\end{equation}

This can be expressed equivalently as:
\begin{equation}
y^2 = 
\begin{cases}
\frac{1}{\lambda}(\lambda g(\mathbf{r}) - w), & \text{if } \lambda g(\mathbf{r}) - w \geq 0, \\
0, & \text{if } \lambda g(\mathbf{r}) - w < 0.
\end{cases}
\label{eq:y_squared_cases}
\end{equation}

Substituting the optimal $y^2$ back into Equation~\eqref{eq:complete_square}, we obtain:
\begin{equation}
-w(g(\mathbf{r}) - y^2) + \frac{\lambda}{2}(g(\mathbf{r}) - y^2)^2 = \frac{1}{2\lambda}\left(z^2 - w^2\right),
\label{eq:substituted}
\end{equation}
where
\begin{equation}
\begin{aligned}
z &= \max\left\{0, w - \lambda(B - F(\mathbf{r}))\right\} \\
&= \max\left\{0, w - \lambda g(\mathbf{r})\right\}.
\end{aligned}
\label{eq:z_definition}
\end{equation}

Therefore, the simplified augmented Lagrangian function, after eliminating $y$, is:
\begin{equation}
\phi(\mathbf{r}, w, \lambda) = \mathcal{L}_{\text{distill}}(\mathbf{r}) + \frac{1}{2\lambda}\left(z^2 - w^2\right),
\label{eq:final_simplified}
\end{equation}
where $z = \max\{0, w - \lambda(B - F(\mathbf{r}))\}$ and $g(\mathbf{r}) = B - F(\mathbf{r})$ represents the constraint satisfaction.


\section{Experiment Details}
\subsection{Experiment Settings}
\label{Appendix: Experiment Settings}

\subsubsection{Training Dataset}

\begin{figure*}[!t]
    \centering
    \includegraphics[width=0.9\textwidth]{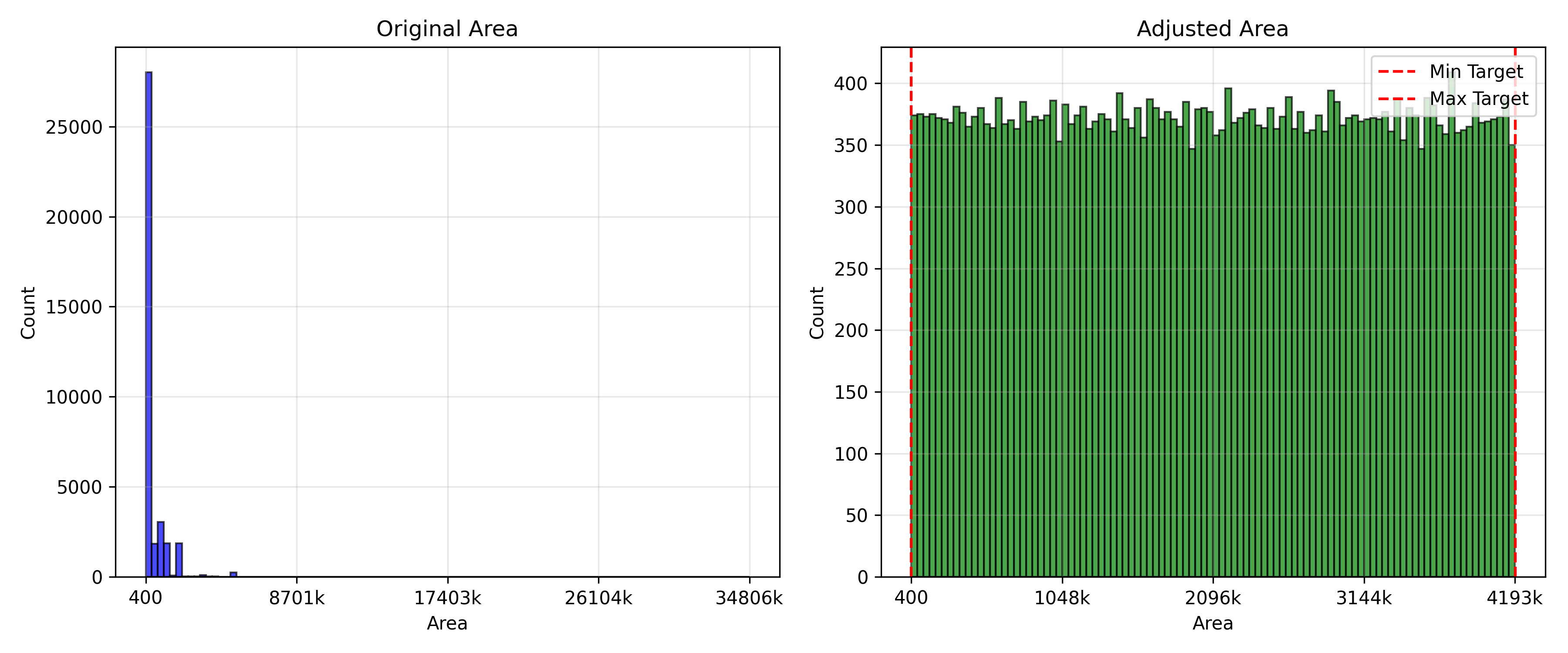}
    \caption{Distribution histogram of image areas in the training dataset before and after applying histogram equalization to balance area diversity. The left panel shows the original distribution heavily concentrated on smaller image areas, while the right panel demonstrates the more balanced distribution after the equalization process.}
    \label{fig:appendix_1}
\end{figure*}

We find that directly training on the original dataset leads to suboptimal performance: \textbf{VisPCO}'s predicted Pareto frontier for high-resolution images concentrates on low computational budgets, diverging from the empirical frontier under high budget regimes. We analyze the training dataset and observe that the distribution of image areas exhibits significant skewness, heavily concentrated on smaller areas, as shown in the left panel of Figure~\ref{fig:appendix_1}. This imbalance is detrimental to learning appropriate pruning ratios, as it leads to poor generalization on high-resolution images. 

To address this issue, we preprocess the training images using histogram equalization to balance the area distribution. Specifically, we divide the image area range into uniform bins and apply stratified sampling to ensure balanced representation across all area intervals. For each bin, we either oversample images (for underrepresented bins) or subsample images (for overrepresented bins) to achieve approximately equal counts per bin. This rebalancing procedure ensures that the training distribution covers the full spectrum of image resolutions uniformly, enabling \textbf{VisPCO} to learn robust pruning configurations for both low and high-resolution images. The left panel of Figure~\ref{fig:appendix_1} shows the original skewed distribution, while the right panel illustrates the balanced distribution after equalization.

\begin{figure}[!t]
    \centering
    \includegraphics[width=0.9\columnwidth]{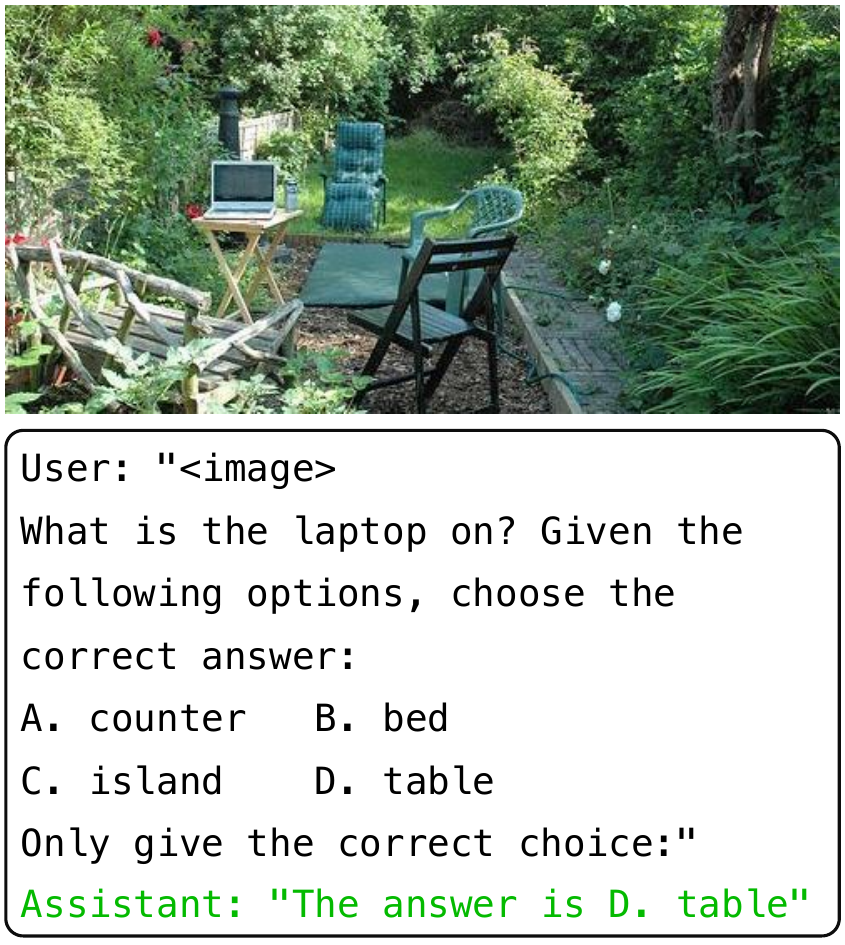}
    \caption{An example evaluation case from the VLMEvalKit benchmark. The figure demonstrates a typical question-answer pair with the corresponding image, showing how the model processes visual and textual inputs to generate responses for evaluation.}
    \label{fig:appendix_2}
    \vspace{-1.5em}
\end{figure}

\subsubsection{Evaluation Datasets}

We utilize the evaluation datasets provided by VLMEvalKit, which includes curated question-answer pairs and images from various vision-language benchmarks. Our evaluation spans three categories of tasks: visual question answering, multimodal reasoning, and chart understanding. For MME, to maintain comparability with other benchmarks, we report the ratio of correct answers to total questions, normalizing the final evaluation results to the range [0, 1]. An example evaluation case is shown in Figure~\ref{fig:appendix_2}.

\textbf{A-OKVQA}~\citep{AOKVQA} is a knowledge-based visual question answering dataset that requires models to leverage external commonsense and world knowledge beyond visual content. It contains 1,145 questions across diverse image types, challenging models to perform reasoning that combines visual understanding with factual knowledge.

\textbf{VizWiz}~\citep{VizWiz} is a visual question answering dataset collected from blind users who took images and asked questions about them. The dataset contains over 4,319 image-question pairs with natural, real-world scenarios, often featuring challenging conditions such as poor image quality, blur, or unusual viewpoints, making it particularly valuable for evaluating model robustness.

\textbf{SEEDBench}~\citep{SeedBench} is a comprehensive benchmark for evaluating multimodal large language models across multiple dimensions. It includes 14,232 multiple-choice questions spanning nine evaluation dimensions including scene understanding, instance identity, spatial relation, and visual reasoning, providing a holistic assessment of model capabilities.

\textbf{MMBench}~\citep{MMBench} (Multimodal Benchmark) is a systematically designed objective benchmark for evaluating various abilities of vision-language models. It covers 20 ability dimensions organized into three categories: perception (e.g., object localization, OCR), reasoning (e.g., social reasoning, physical commonsense), and knowledge (e.g., celebrity recognition, landmark identification).

\textbf{MME}~\citep{MME} (Multi-Modal Evaluation) is a comprehensive evaluation benchmark measuring both perception and cognition abilities. It consists of 14 subtasks including existence, count, position, color, posters, celebrity, scene, landmark, artwork, OCR, commonsense reasoning, numerical calculation, text translation, and code reasoning. We normalize scores to [0, 1] for consistency with other benchmarks.

\textbf{ChartQA}~\citep{ChartQA} focuses on question answering about statistical charts and plots. The dataset contains over 2,000 human-written questions covering bar charts, line plots, and pie charts, requiring models to perform visual reasoning, data extraction, and numerical computation from chart images.

\textbf{OCRBench}~\citep{OCRBench} is a comprehensive benchmark for evaluating optical character recognition and text understanding capabilities in vision-language models. It includes diverse text recognition scenarios such as scene text, handwritten text, document text, and multilingual text, assessing both basic OCR accuracy and text-based reasoning abilities.

\textbf{TextVQA}~\citep{TextVQA} requires models to read and reason about text in images to answer questions. The dataset contains 1,000 images from OpenImages, where answering questions necessitates reading and understanding scene text, making it essential for evaluating text-aware visual reasoning capabilities.

\subsubsection{Pruning Configuration Sampling}

To identify the empirical Pareto frontier that serves as the ground truth for evaluating \textbf{VisPCO}, we employ a comprehensive sampling-based approach. Our methodology consists of three steps: (1) systematically sampling a large number of pruning configurations across the search space, (2) evaluating each configuration's performance across multiple benchmarks and measuring its computational cost in FLOPs, and (3) extracting the Pareto-optimal configurations from the evaluated results.

\textbf{Sampling Strategy.} Our sampling strategy operates at the layer level to capture fine-grained pruning patterns. For a vision-language model with $L$ Transformer layers, we independently sample the visual token retention ratio for each layer from layer 1 to layer $L$. Specifically, the retention ratio $r_i$ for layer $i$ is sampled from the discrete set $\{0.01, 0.06, 0.11, \ldots, 0.96, 0.99\}$, with a uniform step size of 0.05. This granularity balances comprehensive coverage of the configuration space with computational feasibility. For the Qwen2.5-VL-3B model with L=36 layers, this sampling scheme generates a total of 700 distinct pruning configurations spanning diverse computational budgets.

\textbf{Evaluation Protocol.} For each sampled configuration, we perform a complete evaluation to obtain both its performance and computational cost. Performance is measured by averaging accuracy across our eight evaluation benchmarks, providing a comprehensive assessment of model capabilities. Computational cost is calculated using the FLOPs formula derived in Appendix~\ref{Appendix: FLOPs Analysis}, accounting for both the attention mechanism and feed-forward network operations across all layers.

\textbf{Pareto Frontier Extraction.} Given the set of evaluated configurations $\mathcal{C} = \{(p_i, f_i)\}_{i=1}^{N}$, where $p_i \in [0, 1]$ represents the normalized average performance (higher is better) and $f_i$ represents the computational cost in TFLOPs (lower is better) for configuration $i$, we identify the Pareto frontier using the Pareto dominance criterion. Formally, a configuration $(p_i, f_i)$ is said to dominate another configuration $(p_j, f_j)$ if and only if:
\begin{equation}
\begin{aligned}
&p_i \geq p_j \quad \text{and} \quad f_i \leq f_j. \\
\end{aligned}
\end{equation}

The Pareto frontier $\mathcal{P}$ consists of all non-dominated configurations:
\begin{equation}
\begin{aligned}
&\mathcal{P} = \{(p_i, f_i) \in \mathcal{C} \mid \nexists (p_j, f_j) \in \mathcal{C} \\
& \text{such that } (p_j, f_j) \text{ dominates } (p_i, f_i)\}.
\end{aligned}
\end{equation}

\begin{table}[!t]
\centering
\caption{Hyperparameters for main experiments comparing different methods across multiple VLMs.}
\label{tab:hyperparams_main}
\resizebox{\columnwidth}{!}{
\begin{tabular}{lcccccccc}
\toprule
\textbf{Model + Method} & $\lambda$ & $\alpha$ & $\epsilon$ & $\beta$ & $\sigma$ & $T$ & $ \mathrm{lr}$ & $B$ \\
\midrule
Qwen2.5-VL-3B + FastV & 100 & 5 & 0.005 & 0.5 & 10 & 0.1 & 1e-4 & 16 \\
Qwen2.5-VL-3B + SparseVLM & 100 & 5 & 0.005 & 0.5 & 10 & 0.1 & 1e-4 & 16 \\
Qwen2.5-VL-3B + FitPrune & 100 & 5 & 0.005 & 0.5 & 10 & 0.1 & 1e-4 & 16 \\
\midrule
Gemma3-4B + FastV & 1 & 5 & 0.005 & 0.5 & 10 & 0.1 & 5e-4 & 16 \\
Gemma3-4B + SparseVLM & 1 & 5 & 0.005 & 0.5 & 10 & 0.1 & 5e-4 & 16 \\
Gemma3-4B + FitPrune & 1 & 5 & 0.005 & 0.5 & 10 & 0.1 & 5e-4 & 16 \\
\midrule
LLaVA-v1.5-7B + FastV & 100 & 10 & 0.01 & 0.5 & 10 & 0.1 & 5e-5 & 16 \\
LLaVA-v1.5-7B + SparseVLM & 100 & 10 & 0.01 & 0.5 & 10 & 0.1 & 5e-5 & 16 \\
LLaVA-v1.5-7B + FitPrune & 100 & 10 & 0.01 & 0.5 & 10 & 0.1 & 5e-5 & 16 \\
\bottomrule
\end{tabular}
}
\end{table}

\begin{table}[!t]
\centering
\caption{Hyperparameters for ablation studies on different pruning scheduling strategies.}
\label{tab:hyperparams_ablation}
\resizebox{\columnwidth}{!}{
\begin{tabular}{lcccccccc}
\toprule
\textbf{Model + Strategy} & $\lambda$ & $\alpha$ & $\epsilon$ & $\beta$ & $\sigma$ & $T$ & $ \mathrm{lr}$ & $B$ \\
\midrule
Qwen2.5-VL-3B + Linear & 100 & 5 & 0.01 & 0.5 & 1 & 0.1 & 1e-4 & 16 \\
Qwen2.5-VL-3B + Exponential & 100 & 10 & 0.005 & 0.5 & 1 & 0.1 & 5e-5 & 16 \\
Qwen2.5-VL-3B + P-sigmoid & 100 & 5 & 0.01 & 0.5 & 1 & 0.1 & 1e-4 & 16 \\
Qwen2.5-VL-3B + Multi-step & 1 & 10 & 0.05 & 0.5 & 5 & 0.1 & 1e-4 & 16 \\
\bottomrule
\end{tabular}
}
\end{table}

These Pareto-optimal configurations represent the best achievable trade-offs between performance and computational efficiency, forming the empirical frontier against which we evaluate \textbf{VisPCO}'s predictions. This extensive sampling and evaluation process requires significant computational resources (approximately 48+ GPU hours for 700 configurations), highlighting the practical necessity of efficient optimization methods like \textbf{VisPCO}.

\subsubsection{Hyperparameter Settings}

We provide detailed hyperparameter configurations for our experiments in Tables~\ref{tab:hyperparams_main} and~\ref{tab:hyperparams_ablation}. The key hyperparameters and their roles are as follows:

$\lambda$ denotes the penalty parameter in the augmented Lagrangian method, controlling the strength of constraint enforcement. $\epsilon$ is the convergence threshold that determines when the optimization terminates. $\alpha$ and $\beta$ are the update coefficients for the Lagrangian multiplier and penalty parameter, respectively, governing the convergence dynamics. $\sigma$ controls the Gaussian kernel width for continuous relaxation of discrete pruning decisions, with larger values leading to smoother approximations. $T$ is the temperature parameter for the straight-through estimator, balancing between gradient flow and discretization sharpness during training. $ \mathrm{lr}$ denotes the learning rate for the AdamW optimizer, and $B$ indicates the batch size (number of samples per training iteration).

\subsection{More Experiment Results}
\label{Appendix: Different Model Results}
We provide more detailed experimental results in this section. First, we present the results of different pruning methods under various computational budgets with \textbf{VisPCO} in Table~\ref{tab:appendix_1}. Second, we show the results of applying \textbf{VisPCO} to different base VLMs in Table~\ref{tab:appendix_2}. Third, we report the results of \textbf{VisPCO} with different pruning patterns in Table~\ref{tab:appendix_3}.

\begin{figure}[t]
    \centering
    \includegraphics[width=\columnwidth]{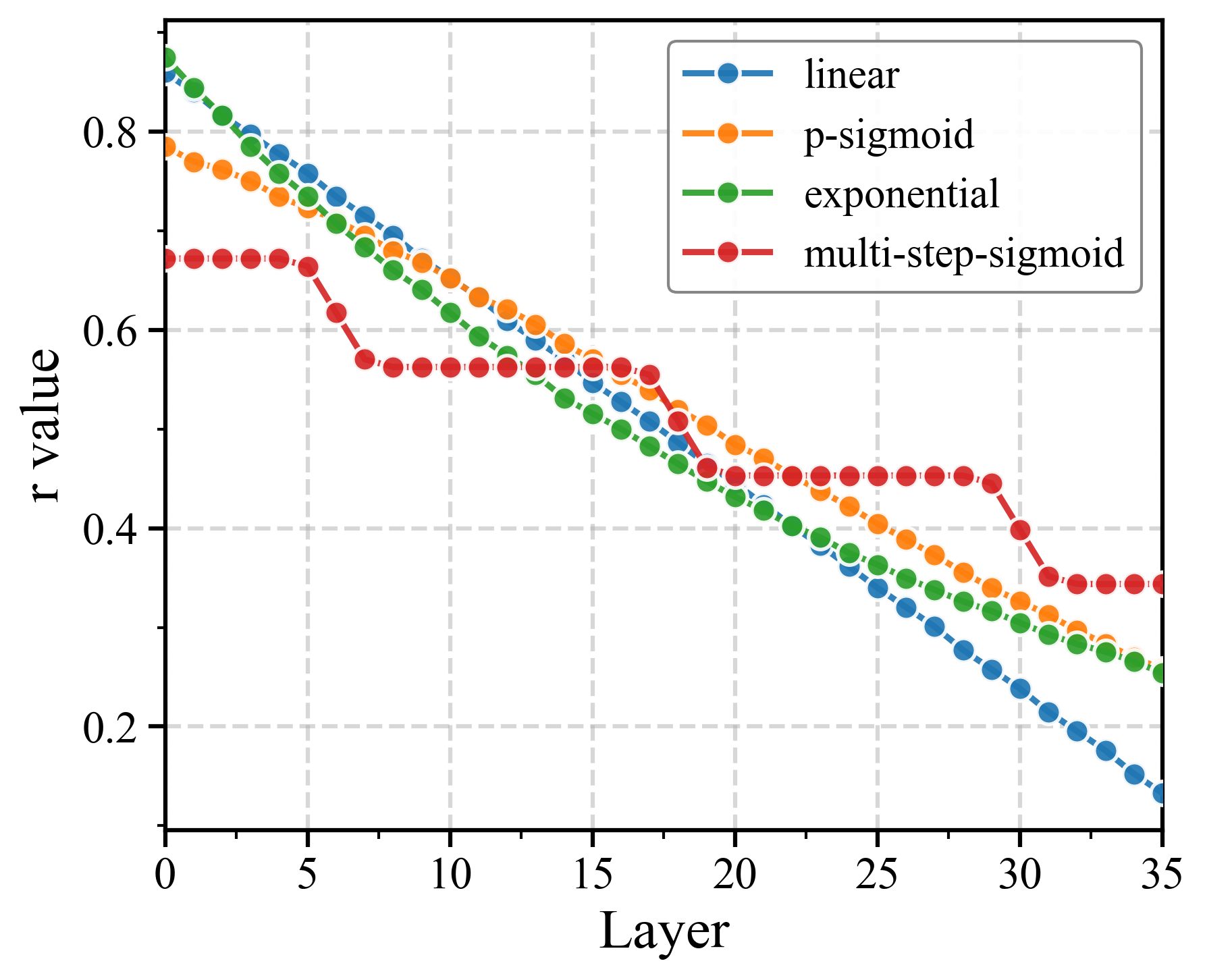}
    \caption{Comparison of layer-wise pruning ratios for different kernel functions under 50\% computational budget. Linear kernel produces gradual transitions across layers, Exponential concentrates pruning in later layers, P-Sigmoid creates smooth S-shaped curves, and Multi-Step generates progressive discrete transitions. The diversity of patterns enables comprehensive exploration of different pruning strategies.}
    \label{fig:appendix_layer_wise_kernel}
\end{figure}

\subsection{Case Studies of Predicted Pruning Configurations}
\label{Appendix: Case Studies}
\begin{figure*}[!t]
    \centering
    \includegraphics[width=0.95\textwidth]{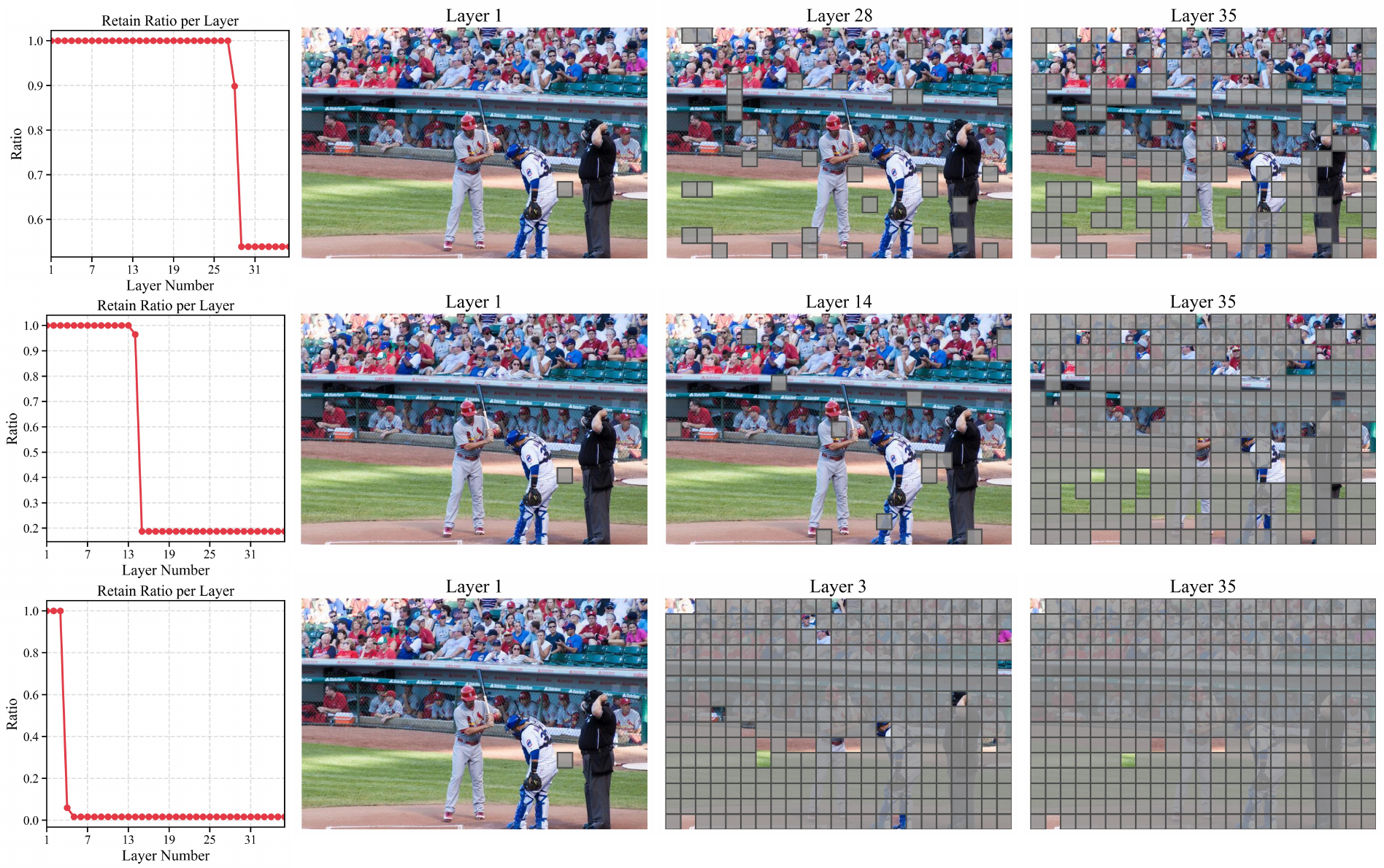}
    \caption{Layer-wise pruning configurations predicted by \textbf{VisPCO} under different computational budgets. The left panel shows the retention ratio curves across layers for budgets ranging from 10\% to 90\%. The right panel visualizes the actual visual token retention at selected layers (from top to bottom: 90\%, 50\%, 10\% budgets), demonstrating how aggressive pruning (lower budgets) leads to earlier and more extensive token removal.}
    \label{fig:appendix_case_study}
\end{figure*}

We present case studies of pruning configurations predicted by \textbf{VisPCO} on Qwen2.5-VL-3B to provide insights into its behavior under different computational budgets. Figure~\ref{fig:appendix_case_study} illustrates the layer-wise pruning curves predicted by \textbf{VisPCO} under various budget constraints, along with the corresponding visual token retention patterns at different layers. These visualizations reveal how \textbf{VisPCO} adaptively adjusts its pruning strategy in response to varying resource constraints.

The visualization reveals several key observations. First, as the computational budget becomes more constrained, \textbf{VisPCO} adopts increasingly aggressive pruning strategies, with pruning occurring earlier in the network and achieving lower retention ratios. This demonstrates the model's ability to adaptively allocate computational resources based on budget constraints. Second, the predicted configurations exhibit smooth transitions across layers, validating the effectiveness of our continuous relaxation approach.

Additionally, Figure~\ref{fig:appendix_layer_wise_kernel} presents a comparison of different kernel functions (Linear, Exponential, P-Sigmoid, Multi-Step) for multi-layer pruning under a 50\% computational budget. The layer-wise pruning ratios reveal distinct patterns: linear kernels produce gradual transitions, exponential kernels concentrate pruning in later layers, p-sigmoid kernels create smooth S-shaped curves, and multi-step kernels generate progressive discrete transitions. These diverse patterns enable \textbf{VisPCO} to explore different trade-offs.

\begin{table*}[!t]
\centering
\caption{Detailed comparison of pruning methods with and without \textbf{VisPCO} across eight vision-language benchmarks under different computational budgets. Results without \textbf{VisPCO} are averaged over multiple randomly sampled configurations that satisfy the budget constraint, with standard deviations reported in parentheses.}
\label{tab:appendix_1}
\resizebox{\textwidth}{!}{

\begin{tabular}{l|cccccccc|c}
\noalign{
  \hrule height 1pt
}
\textbf{Method} & \textbf{AOKVQA} & \textbf{VizWiz} & \textbf{SEED} & \textbf{MMB} & \textbf{MME}$^{\dagger}$ & \textbf{ChartQA} & \textbf{OCRB} & \textbf{TextVQA} & \textbf{Avg (\%)}\\
\noalign{
  \hrule height 1pt
}
\rowcolor{gray!20}
\multicolumn{10}{c}{\textit{Upper Bound, 100\% Budget, $\sim$3.56 TFLOPs}} \\
Qwen2.5VL-3B & $90.2$ & $75.1$ & $75.6$ & $79.8$ & $84.2$ & $64.1$ & $74.6$ & $81.3$ & $78.1$ \\
\noalign{
  \hrule height 1pt
}
\rowcolor{gray!20}
\multicolumn{10}{c}{\textit{Reduce FLOPs Budget to 90\%, $\sim$3.20 TFLOPs}} \\
$\llcorner$ FastV & $88.2 \pm 0.4$ & $72.9 \pm 0.9$ & $72.4 \pm 0.9$ & $76.4 \pm 0.5$ & $81.3 \pm 0.5$ & $62.2 \pm 0.8$ & $71.6 \pm 0.7$ & $79.1 \pm 0.6$ & $75.5 \pm 0.7$\\
\;\;+ VisPCO & $88.4$ & $73.8$ & $73.2$ & $76.9$ & $81.7$ & $62.9$ & $72.3$ & $79.5$ & $76.1$\\
$\llcorner$ SparseVLM & $88.5 \pm 0.3$ & $73.1 \pm 0.5$ & $73.4 \pm 0.4$ & $76.9 \pm 0.6$ & $82.1 \pm 0.3$ & $62.2 \pm 0.7$ & $71.9 \pm 0.6$ & $79.5 \pm 0.6$ & $76.0 \pm 0.5$\\
\;\;+ VisPCO & $88.6$ & $73.5$ & $73.8$ & $77.5$ & $82.4$ & $62.9$ & $72.5$ & $80.0$ & $76.4$\\
$\llcorner$ FitPrune & $89.1 \pm 0.5$ & $73.9 \pm 0.4$ & $74.2 \pm 0.5$ & $77.6 \pm 0.4$ & $82.5 \pm 0.6$ & $63.1 \pm 0.6$ & $72.5 \pm 0.5$ & $79.9 \pm 0.3$ & $76.2 \pm 0.5$\\
\;+\; VisPCO & $89.6$ & $74.1$ & $74.6$ & $77.9$ & $82.8$ & $63.5$ & $72.9$ & $81.2$ & $77.1$\\

\noalign{
  \hrule height 1pt
}
\rowcolor{gray!20}
\multicolumn{10}{c}{\textit{Reduce FLOPs Budget to 80\%, $\sim$2.84 TFLOPs}} \\
$\llcorner$ FastV & $87.0 \pm 1.6$ & $71.6 \pm 2.1$ & $71.2 \pm 2.1$ & $75.1 \pm 1.8$ & $80.1 \pm 1.6$ & $61.0 \pm 2.2$ & $70.2 \pm 1.9$ & $77.7 \pm 2.2$ & $74.2 \pm 1.9$\\
\;\;+ VisPCO & $88.3$ & $73.7$ & $73.1$ & $76.9$ & $81.6$ & $62.8$ & $72.0$ & $79.4$ & $75.7$\\
$\llcorner$ SparseVLM & $87.1 \pm 1.8$ & $71.7 \pm 2.4$ & $71.4 \pm 2.1$ & $75.3 \pm 2.1$ & $80.2 \pm 2.2$ & $61.6 \pm 2.2$ & $70.5 \pm 2.1$ & $78.3 \pm 2.0$ & $74.5 \pm 2.1$\\
\;\;+ VisPCO & $88.4$ & $73.9$ & $73.5$ & $77.4$ & $82.1$ & $62.6$ & $72.4$ & $79.6$ & $76.2$\\
$\llcorner$ FitPrune & $87.3 \pm 2.2$ & $72.3 \pm 2.1$ & $72.5 \pm 2.3$ & $75.8 \pm 2.2$ & $80.6 \pm 2.2$ & $61.6 \pm 2.5$ & $70.6 \pm 2.5$ & $78.4 \pm 1.6$ & $74.9 \pm 2.2$\\
\;+\; VisPCO & $89.3$ & $73.8$ & $74.3$ & $77.4$ & $82.6$ & $63.3$ & $72.5$ & $79.9$ & $76.6$\\
\noalign{
  \hrule height 1pt
}
\rowcolor{gray!20}
\multicolumn{10}{c}{\textit{Reduce FLOPs Budget to 70\%, $\sim$2.50 TFLOPs}} \\
$\llcorner$ FastV & $83.9 \pm 4.5$ & $69.2 \pm 4.4$ & $68.4 \pm 4.8$ & $73.1 \pm 3.8$ & $76.2 \pm 4.5$ & $59.3 \pm 4.3$ & $66.2 \pm 4.9$ & $75.5 \pm 4.1$ & $71.5 \pm 4.4$\\
\;\;+ VisPCO & $88.0$ & $73.4$ & $72.8$ & $76.7$ & $80.9$ & $62.4$ & $71.1$ & $78.9$ & $75.5$\\
$\llcorner$ SparseVLM & $84.1 \pm 4.7$ & $69.4 \pm 4.6$ & $68.5 \pm 5.1$ & $73.3 \pm 4.0$ & $76.3 \pm 4.5$ & $59.5 \pm 4.5$ & $66.3 \pm 5.0$ & $75.7 \pm 4.2$ & $71.6 \pm 4.6$\\
\;\;+ VisPCO & $88.1$ & $73.6$ & $72.9$ & $76.7$ & $81.0$ & $62.5$ & $71.1$ & $79.1$ & $75.6$\\
$\llcorner$ FitPrune & $84.2 \pm 4.6$ & $69.5 \pm 4.7$ & $68.6 \pm 5.2$ & $73.5 \pm 4.1$ & $76.6 \pm 4.5$ & $59.4 \pm 4.7$ & $66.5 \pm 5.2$ & $75.8 \pm 4.3$ & $71.8 \pm 4.7$\\
\;+\; VisPCO & $88.2$ & $73.7$ & $73.1$ & $76.7$ & $81.1$ & $62.8$ & $71.0$ & $79.3$ & $75.7$\\
\noalign{
  \hrule height 1pt
}
\rowcolor{gray!20}
\multicolumn{10}{c}{\textit{Reduce FLOPs Budget to 60\%, $\sim$2.14 TFLOPs}} \\
$\llcorner$ FastV & $80.2 \pm 5.9$ & $68.4 \pm 5.4$ & $66.6 \pm 6.8$ & $69.4 \pm 5.3$ & $74.5 \pm 5.5$ & $56.4 \pm 5.3$ & $65.2 \pm 5.1$ & $71.7 \pm 5.1$ & $69.2 \pm 5.4$\\
\;\;+ VisPCO & $86.0$ & $71.4$ & $70.8$ & $74.7$ & $78.9$ & $60.4$ & $69.1$ & $76.9$ & $73.5$\\
$\llcorner$ SparseVLM & $80.3 \pm 6.2$ & $68.5 \pm 5.5$ & $66.7 \pm 7.1$ & $69.6 \pm 5.6$ & $74.8 \pm 5.8$ & $56.4 \pm 5.6$ & $65.7 \pm 5.5$ & $71.8 \pm 5.3$ & $69.4 \pm 5.7$\\
\;\;+ VisPCO & $86.2$ & $71.6$ & $70.9$ & $74.9$ & $79.2$ & $60.6$ & $69.3$ & $77.1$ & $73.7$\\
$\llcorner$ FitPrune & $80.4 \pm 6.1$ & $68.6 \pm 5.4$ & $66.8 \pm 7.0$ & $69.8 \pm 5.8$ & $74.9 \pm 5.9$ & $56.7 \pm 5.9$ & $65.9 \pm 5.7$ & $71.9 \pm 5.3$ & $69.5 \pm 5.8$\\
\;\;+ VisPCO & $86.3$ & $71.7$ & $71.0$ & $75.1$ & $79.4$ & $60.9$ & $69.6$ & $77.1$ & $73.9$\\
\noalign{
  \hrule height 1pt
}
\rowcolor{gray!20}
\multicolumn{10}{c}{\textit{Reduce FLOPs Budget to 50\%, $\sim$1.78 TFLOPs}} \\
$\llcorner$ FastV& $74.7 \pm 10.1$ & $60.3 \pm 9.6$ & $61.5 \pm 8.1$ & $62.4 \pm 9.3$ & $68.8 \pm 9.1$ & $51.6 \pm 9.9$ & $59.2 \pm 9.1$ & $65.9 \pm 10.8$ & $63.1 \pm 9.5$\\
\;\;+ VisPCO & $84.8$ & $69.4$ & $67.6$ & $71.2$ & $77.1$ & $58.1$ & $67.8$ & $75.9$ & $71.5$\\
$\llcorner$ SparseVLM & $ 75.9 \pm 9.8$ & $62.6 \pm 8.2$ & $63.1 \pm 7.2$ & $63.9 \pm 8.6$ & $69.9 \pm 8.2$ & $51.9 \pm 8.3$ & $62.4 \pm 6.9$ & $66.6 \pm 9.8$ & $64.5 \pm 8.4$\\
\;\;+ VisPCO & $85.2$ & $69.0$ & $68.1$ & $71.9$ & $77.6$ & $58.4$ & $67.9$ & $76.3$ & $71.8$\\
$\llcorner$ FitPrune & $ 77.1 \pm 8.7$ & $63.4 \pm 7.7$ & $63.9 \pm 6.8$ & $64.5 \pm 8.2$ & $70.8 \pm 7.9$ & $52.8 \pm 8.1$ & $63.3 \pm 6.4$ & $67.6 \pm 9.4$ & $65.4 \pm 7.9$\\
\;\;+ VisPCO & $85.9$ & $69.4$ & $68.4$ & $72.4$ & $77.9$ & $58.8$ & $68.2$ & $76.6$ & $72.2$\\
\noalign{
  \hrule height 1pt
}
\rowcolor{gray!20}
\multicolumn{10}{c}{\textit{Reduce FLOPs Budget to 40\%, $\sim$1.42 TFLOPs}} \\
$\llcorner$ FastV & $65.6 \pm 12.2$ & $50.3 \pm 11.4$ & $52.6 \pm 10.2$ & $53.3 \pm 11.3$ & $61.4 \pm 11.2$ & $42.5 \pm 11.7$ & $51.4 \pm 11.2$ & $57.5 \pm 12.7$ & $54.3 \pm 11.5$\\
\;\;+ VisPCO & $77.6$ & $61.7$ & $62.6$ & $64.4$ & $72.6$ & $54.1$ & $62.5$ & $69.9$ & $65.7$\\
$\llcorner$ SparseVLM & $66.8 \pm 12.6$ & $50.9 \pm 11.8$ & $53.2 \pm 10.9$ & $53.5 \pm 11.8$ & $62.3 \pm 11.3$ & $42.9 \pm 11.9$ & $51.8 \pm 11.3$ & $58.1 \pm 12.9$ & $54.9 \pm 11.8$\\
\;\;+ VisPCO & $77.9$ & $61.9$ & $62.7$ & $64.9$ & $72.9$ & $54.4$ & $62.6$ & $70.2$ & $65.9$\\
$\llcorner$ FitPrune & $66.9 \pm 12.5$ & $50.7 \pm 12.2$ & $53.0 \pm 11.1$ & $53.7 \pm 11.9$ & $62.7 \pm 11.5$ & $42.8 \pm 12.2$ & $51.7 \pm 11.5$ & $58.3 \pm 12.6$ & $55.0 \pm 11.9$\\
\;\;+ VisPCO & $78.4$ & $62.2$ & $62.8$ & $65.0$ & $73.1$ & $54.6$ & $62.6$ & $70.4$ & $66.1$\\
\noalign{
  \hrule height 1pt
}
\rowcolor{gray!20}
\multicolumn{10}{c}{\textit{Reduce FLOPs Budget to 30\%, $\sim$1.06 TFLOPs}} \\
$\llcorner$ FastV & $62.5 \pm 8.1$ & $46.4 \pm 8.2$ & $48.7 \pm 7.6$ & $49.4 \pm 7.5$ & $57.2 \pm 8.9$ & $38.8 \pm 8.1$ & $46.7 \pm 8.3$ & $53.6 \pm 9.1$ & $50.4 \pm 8.2$\\
\;\;+ VisPCO & $70.2$ & $54.6$ & $55.5$ & $54.5$ & $65.7$ & $46.4$ & $54.4$ & $62.8$ & $58.0$\\
$\llcorner$ SparseVLM & $62.6 \pm 8.3$ & $46.5 \pm 8.3$ & $48.9 \pm 7.7$ & $49.5 \pm 7.7$ & $57.4 \pm 9.0$ & $38.9 \pm 8.3$ & $46.9 \pm 8.5$ & $53.8 \pm 9.3$ & $50.6 \pm 8.4$\\
\;\;+ VisPCO & $70.5$ & $54.8$ & $55.9$ & $54.8$ & $65.8$ & $46.9$ & $55.1$ & $63.4$ & $58.4$\\
$\llcorner$ FitPrune & $62.7 \pm 8.2$ & $46.7 \pm 8.2$ & $49.0 \pm 7.6$ & $49.7 \pm 7.9$ & $57.3 \pm 9.1$ & $39.1 \pm 8.1$ & $46.7 \pm 8.6$ & $53.9 \pm 9.4$ & $50.6 \pm 8.4$\\
\;\;+ VisPCO & $70.6$ & $54.9$ & $56.0$ & $54.7$ & $65.8$ & $47.1$ & $55.3$ & $63.1$ & $58.4$\\
\noalign{
  \hrule height 1pt
}
\rowcolor{gray!20}
\multicolumn{10}{c}{\textit{Reduce FLOPs Budget to 20\%, $\sim$0.72 TFLOPs}} \\
$\llcorner$ FastV & $42.5 \pm 4.1$ & $39.4 \pm 4.2$ & $46.7 \pm 3.6$ & $39.4 \pm 4.5$ & $47.2 \pm 4.9$ & $34.8 \pm 4.1$ & $12.7 \pm 5.3$ & $43.6 \pm 5.1$ & $38.3 \pm 4.5$\\
\;\;+ VisPCO & $46.6$ & $43.1$ & $50.1$ & $43.9$ & $51.2$ & $38.9$ & $17.8$ & $48.5$ & $42.5$\\
$\llcorner$ SparseVLM & $42.6 \pm 4.2$ & $39.5 \pm 4.3$ & $46.9 \pm 3.7$ & $39.6 \pm 4.6$ & $47.4 \pm 5.0$ & $34.9 \pm 4.2$ & $12.8 \pm 5.4$ & $43.7 \pm 5.3$ & $38.4 \pm 4.6$\\
\;\;+ VisPCO & $46.7$ & $43.2$ & $50.2$ & $43.9$ & $51.3$ & $39.1$ & $17.9$ & $48.6$ & $42.6$\\
$\llcorner$ FitPrune & $42.5 \pm 4.3$ & $39.4 \pm 4.4$ & $46.7 \pm 3.9$ & $39.7 \pm 4.7$ & $47.6 \pm 5.1$ & $34.7 \pm 4.1$ & $12.6 \pm 5.2$ & $43.8 \pm 5.4$ & $38.3 \pm 4.6$\\
\;\;+ VisPCO & $46.8$ & $43.3$ & $50.4$ & $44.1$ & $51.4$ & $39.2$ & $18.0$ & $48.7$ & $42.7$\\
\noalign{
  \hrule height 1pt
}
\rowcolor{gray!20}
\multicolumn{10}{c}{\textit{Reduce FLOPs Budget to 10\%, $\sim$0.36 TFLOPs}} \\
$\llcorner$ FastV & $ 33.3 \pm 2.3$ & $30.4 \pm 1.6$ & $44.5 \pm 2.7$ & $33.0 \pm 2.5$ & $39.7 \pm 1.4$ & $29.8 \pm 4.1$ & $8.3 \pm 2.1$ & $33.7 \pm 2.8$ & $31.6 \pm 2.4$\\
\;\;+ VisPCO & $35.5$ & $31.7$ & $46.9$ & $35.5$ & $40.1$ & $33.2$ & $10.1$ & $36.1$ & $33.6$\\
$\llcorner$ SparseVLM & $ 33.6 \pm 2.1$ & $31.2 \pm 1.3$ & $44.9 \pm 2.5$ & $33.9 \pm 2.3$ & $40.3 \pm 1.1$ & $30.5 \pm 3.7$ & $9.1 \pm 2.0$ & $34.4 \pm 2.2$ & $32.2 \pm 2.2 $\\
\;\;+ VisPCO & $35.5$ & $31.5$ & $47.1$ & $35.8$ & $40.4$ & $33.3$ & $10.2$ & $36.3$ & $33.8$\\
$\llcorner$ FitPrune & $ 33.8 \pm 2.1$ & $31.5 \pm 1.1$ & $45.3 \pm 2.4$ & $34.2 \pm 2.2$ & $40.6 \pm 1.0$ & $30.9 \pm 3.5$ & $9.6 \pm 1.9$ & $34.6 \pm 2.1$ & $32.6 \pm 2.0 $\\
\;\;+ VisPCO & $35.6$ & $31.6$ & $47.3$ & $35.8$ & $40.9$ & $33.5$ & $10.4$ & $36.4$ & $33.9$\\
\noalign{
  \hrule height 1pt
}
\end{tabular}}
\vspace{-1em}
\end{table*}

\begin{table*}[!t]
\centering
\caption{Performance of \textbf{VisPCO} applied to different base vision-language models across eight benchmarks under various computational budgets. The results demonstrate the generalizability and effectiveness of \textbf{VisPCO} across different model architectures and sizes.}
\label{tab:appendix_2}
\resizebox{\textwidth}{!}{

\begin{tabular}{l|cccccccc|c}
\noalign{
  \hrule height 1pt
}
\textbf{Method} & \textbf{AOKVQA} & \textbf{VizWiz} & \textbf{SEED} & \textbf{MMB} & \textbf{MME}$^{\dagger}$ & \textbf{ChartQA} & \textbf{OCRB} & \textbf{TextVQA} & \textbf{Avg (\%)}\\
\noalign{
  \hrule height 1pt
}
\rowcolor{gray!20}
\multicolumn{10}{c}{\textit{Upper Bound, 100\% Budget}} \\
LLaVA-7B & $72.3$ & $93.1$ & $52.1$ & $48.2$ & $50.4$ & $42.3$ & $64.3$ & $88.2$ & $63.9$ \\
Gemma3-4B & $80.1$ & $61.2$ & $69.9$ & $72.3$ & $79.3$ & $53.8$ & $64.2$ & $70.3$ & $68.9$ \\
\noalign{
  \hrule height 1pt
}
\rowcolor{gray!20}
\multicolumn{10}{c}{\textit{Reduce FLOPs Budget to 90\%}} \\
LLaVA-7B & $71.3 \pm 0.5$ & $91.5 \pm 0.8$ & $50.3 \pm 0.6$ & $47.3 \pm 0.6$ & $48.4 \pm 0.8$ & $40.9 \pm 0.7$ & $62.8 \pm 0.7$ & $86.2 \pm 0.5$ & $62.3 \pm 0.7$\\
\;+ VisPCO & $71.8$ & $92.3$ & $50.8$ & $47.7$ & $49.2$ & $41.5$ & $63.5$ & $86.5$ & $62.9$\\
Gemma3-4B & $78.8 \pm 0.5$ & $59.8 \pm 0.9$ & $67.7 \pm 0.7$ & $70.5 \pm 0.6$ & $77.6 \pm 0.9$ & $51.8 \pm 0.8$ & $62.3 \pm 0.7$ & $58.8 \pm 0.5$ & $65.9 \pm 0.7$\\
\;+ VisPCO & $79.3$ & $60.5$ & $68.4$ & $71.1$ & $78.5$ & $52.5$ & $62.9$ & $59.2$ & $66.6$\\
\noalign{
  \hrule height 1pt
}
\rowcolor{gray!20}
\multicolumn{10}{c}{\textit{Reduce FLOPs Budget to 80\%}} \\
LLaVA-7B & $70.1 \pm 1.6$ & $89.7 \pm 1.7$ & $48.7 \pm 1.3$ & $45.4 \pm 1.3$ & $46.2 \pm 1.7$ & $38.5 \pm 1.6$ & $60.6 \pm 1.7$ & $84.1 \pm 1.6$ & $60.2 \pm 1.9$\\
\;+ VisPCO & $71.7$ & $91.4$ & $49.9$ & $46.7$ & $47.6$ & $40.0$ & $62.3$ & $85.6$ & $61.9$\\
Gemma3-4B & $76.6 \pm 1.4$ & $57.9 \pm 1.7$ & $65.6 \pm 1.6$ & $68.4 \pm 1.6$ & $75.3 \pm 1.8$ & $49.8 \pm 1.8$ & $60.2 \pm 1.6$ & $56.6 \pm 1.5$ & $63.8 \pm 1.6$\\
\;+ VisPCO & $77.8$ & $59.5$ & $67.2$ & $69.9$ & $77.0$ & $51.5$ & $61.8$ & $57.9$ & $65.3$\\
\noalign{
  \hrule height 1pt
}
\rowcolor{gray!20}
\multicolumn{10}{c}{\textit{Reduce FLOPs Budget to 70\%}} \\
LLaVA-7B & $66.2 \pm 4.7$ & $85.7 \pm 4.5$ & $44.6 \pm 4.3$ & $41.3 \pm 4.4$ & $42.3 \pm 4.6$ & $34.6 \pm 4.7$ & $56.3 \pm 4.5$ & $80.3 \pm 4.1$ & $56.4 \pm 4.5$\\
\;+ VisPCO & $70.7$ & $89.3$ & $48.9$ & $45.7$ & $46.6$ & $39.1$ & $60.7$ & $84.4$ & $60.7$\\
Gemma3-4B & $70.2 \pm 5.1$ & $51.7 \pm 5.4$ & $59.5 \pm 5.5$ & $62.3 \pm 5.2$ & $69.1 \pm 4.7$ & $43.7 \pm 5.4$ & $54.3 \pm 5.4$ & $50.6 \pm 6.2$ & $57.7 \pm 5.4$\\
\;+ VisPCO & $75.3$ & $56.9$ & $65.1$ & $67.5$ & $73.5$ & $58.9$ & $59.7$ & $56.7$ & $64.2$\\
\noalign{
  \hrule height 1pt
}
\rowcolor{gray!20}
\multicolumn{10}{c}{\textit{Reduce FLOPs Budget to 60\%}} \\
LLaVA-7B & $62.1 \pm 6.6$ & $82.9 \pm 6.2$ & $41.7 \pm 6.4$ & $38.4 \pm 5.6$ & $38.1 \pm 7.0$ & $31.6 \pm 6.5$ & $53.5 \pm 6.1$ & $76.1 \pm 6.3$ & $53.1 \pm 6.3$\\
\;+ VisPCO & $68.1$ & $88.3$ & $47.5$ & $43.6$ & $44.1$ & $37.0$ & $59.6$ & $81.3$ & $58.7$\\
Gemma3-4B & $54.1 \pm 6.1$ & $46.6 \pm 6.3$ & $54.4 \pm 6.6$ & $57.2 \pm 7.3$ & $65.1 \pm 6.8$ & $38.6 \pm 6.2$ & $49.4 \pm 6.5$ & $45.6 \pm 7.2$ & $53.1 \pm 6.6$\\
\;+ VisPCO & $60.2$ & $52.9$ & $60.9$ & $64.1$ & $71.1$ & $44.3$ & $54.9$ & $51.3$ & $57.5$\\
\noalign{
  \hrule height 1pt
}
\rowcolor{gray!20}
\multicolumn{10}{c}{\textit{Reduce FLOPs Budget to 50\%}} \\
LLaVA-7B & $50.3 \pm 10.7$ & $70.8 \pm 10.3$ & $30.8 \pm 11.3$ & $26.5 \pm 11.5$ & $26.1 \pm 11.4$ & $21.3 \pm 10.6$ & $41.2 \pm 11.6$ & $64.2 \pm 11.4$ & $41.4 \pm 11.1$\\
\;+ VisPCO & $60.8$ & $80.5$ & $42.1$ & $38.0$ & $37.4$ & $31.6$ & $52.8$ & $75.2$ & $52.3$\\
Gemma3-4B & $41.2 \pm 12.4$ & $33.5 \pm 12.6$ & $41.8 \pm 11.5$ & $44.7 \pm 12.9$ & $52.3 \pm 11.8$ & $25.7 \pm 12.3$ & $37.6 \pm 12.7$ & $33.6 \pm 12.3$ & $38.8 \pm 12.3$\\
\;+ VisPCO & $53.4$ & $45.9$ & $53.1$ & $57.3$ & $62.8$ & $37.9$ & $50.3$ & $45.5$ & $50.8$\\
\noalign{
  \hrule height 1pt
}
\rowcolor{gray!20}
\multicolumn{10}{c}{\textit{Reduce FLOPs Budget to 40\%}} \\
LLaVA-7B & $48.4 \pm 8.6$ & $68.5 \pm 8.4$ & $29.7 \pm 8.8$ & $24.1 \pm 9.4$ & $24.2 \pm 9.8$ & $19.5 \pm 8.6$ & $39.3 \pm 9.4$ & $62.7 \pm 9.9$ & $39.6 \pm 9.1$\\
\;+ VisPCO & $56.8$ & $76.9$ & $38.4$ & $33.5$ & $34.0$ & $27.9$ & $48.7$ & $71.7$ & $48.5$\\
Gemma3-4B & $39.4 \pm 8.1$ & $31.1 \pm 9.4$ & $39.8 \pm 9.7$ & $42.8 \pm 8.9$ & $50.4 \pm 9.5$ & $23.6 \pm 8.4$ & $35.8 \pm 8.9$ & $31.7 \pm 8.3$ & $36.8 \pm 8.9$\\
\;+ VisPCO & $47.5$ & $40.1$ & $48.7$ & $51.5$ & $59.9$ & $31.4$ & $44.2$ & $39.9$ & $45.4$\\
\noalign{
  \hrule height 1pt
}
\rowcolor{gray!20}
\multicolumn{10}{c}{\textit{Reduce FLOPs Budget to 30\%}} \\
LLaVA-7B & $41.3 \pm 6.7$ & $61.6 \pm 6.1$ & $22.8 \pm 6.5$ & $17.2 \pm 6.5$ & $17.4 \pm 6.9$ & $14.6 \pm 6.7$ & $36.4 \pm 6.3$ & $60.8 \pm 8.2$ & $34.0 \pm 6.7$\\
\;+ VisPCO & $47.8$ & $67.7$ & $29.2$ & $23.6$ & $24.2$ & $20.9$ & $42.5$ & $68.9$ & $40.6$\\
Gemma3-4B & $31.4 \pm 6.1$ & $24.1 \pm 6.4$ & $31.8 \pm 6.7$ & $36.8 \pm 6.9$ & $44.4 \pm 6.5$ & $20.6 \pm 6.4$ & $31.8 \pm 6.9$ & $25.7 \pm 6.3$ & $30.8 \pm 6.5$\\
\;+ VisPCO & $37.5$ & $30.1$ & $38.5$ & $43.5$ & $50.2$ & $27.0$ & $38.3$ & $31.9$ & $37.1$\\
\noalign{
  \hrule height 1pt
}
\rowcolor{gray!20}
\multicolumn{10}{c}{\textit{Reduce FLOPs Budget to 20\%}} \\
LLaVA-7B & $40.8 \pm 2.8$ & $60.9 \pm 2.2$ & $21.7 \pm 2.6$ & $17.1 \pm 2.9$ & $16.9 \pm 3.1$ & $13.8 \pm 2.5$ & $34.6 \pm 2.5$ & $60.1 \pm 2.3$ & $37.2 \pm 2.6$\\
\;+ VisPCO & $43.6$ & $63.1$ & $24.3$ & $19.9$ & $19.8$ & $16.3$ & $37.1$ & $62.4$ & $39.8$\\
Gemma3-4B & $31.2 \pm 2.2$ & $23.8 \pm 2.4$ & $31.9 \pm 2.2$ & $36.4 \pm 2.5$ & $44.6 \pm 2.6$ & $20.3 \pm 2.6$ & $31.7 \pm 2.3$ & $25.6 \pm 2.3$ & $30.7 \pm 2.4$\\
\;+ VisPCO & $33.4$ & $26.2$ & $34.0$ & $38.9$ & $47.2$ & $22.9$ & $34.0$ & $27.9$ & $33.1$\\
\noalign{
  \hrule height 1pt
}
\rowcolor{gray!20}
\multicolumn{10}{c}{\textit{Reduce FLOPs Budget to 10\%}} \\
LLaVA-7B & $40.1 \pm 0.7$ & $60.8 \pm 0.8$ & $21.4 \pm 0.6$ & $17.2 \pm 0.4$ & $16.7 \pm 0.2$ & $13.5 \pm 0.3$ & $34.3 \pm 0.9$ & $60.2 \pm 0.5$ & $36.8 \pm 0.6$\\
\;+ VisPCO & $40.7$ & $61.5$ & $22.0$ & $17.6$ & $16.8$ & $13.8$ & $35.0$ & $60.6$ & $33.5$\\
Gemma3-4B & $30.1 \pm 0.3$ & $22.8 \pm 0.5$ & $27.4 \pm 0.6$ & $33.3 \pm 0.8$ & $41.5 \pm 0.7$ & $19.2 \pm 0.6$ & $30.6 \pm 0.6$ & $23.5 \pm 0.4$ & $28.6 \pm 0.6$\\
\;+ VisPCO & $30.4$ & $23.2$ & $27.9$ & $34.1$ & $42.2$ & $19.7$ & $31.2$ & $23.9$ & $29.1$\\
\noalign{
  \hrule height 1pt
}
\end{tabular}}
\vspace{-1em}
\end{table*}

\begin{table*}[!t]
\centering
\caption{Performance comparison of \textbf{VisPCO} with different pruning patterns across vision-language benchmarks. The table shows detailed results for single-layer pruning and various multi-layer pruning strategies including Linear, Exponential, P-Sigmoid, and Multi-Step patterns.}
\label{tab:appendix_3}
\resizebox{\textwidth}{!}{

\begin{tabular}{l|ccccccccc|c}
\noalign{
  \hrule height 1pt
}
\textbf{Method} & \textbf{Kernels} & \textbf{AOKVQA} & \textbf{VizWiz} & \textbf{SEED} & \textbf{MMB} & \textbf{MME}$^{\dagger}$ & \textbf{ChartQA} & \textbf{OCRB} & \textbf{TextVQA} & \textbf{Avg (\%)}\\
\noalign{
  \hrule height 1pt
}
\rowcolor{gray!20}
\multicolumn{11}{c}{\textit{Upper Bound, 100\% Budget, $\sim$3.56 TFLOPs}} \\
Qwen2.5VL-3B & - & $90.2$ & $75.1$ & $75.6$ & $79.8$ & $84.2$ & $64.1$ & $74.6$ & $81.3$ & $78.1$ \\
\noalign{
  \hrule height 1pt
}
\rowcolor{gray!20}
\multicolumn{11}{c}{\textit{Reduce FLOPs Budget to 90\%, $\sim$3.20 TFLOPs}} \\
Single-Layer & - & $88.4$ & $73.8$ & $73.2$ & $76.9$ & $81.7$ & $62.9$ & $72.3$ & $79.5$ & $76.1$\\
\noalign{
  \hrule height 0.5pt
}
& Linear & $88.2$ & $73.1$ & $72.5$ & $76.4$ & $81.3$ & $61.8$ & $72.0$ & $79.1$ & $75.6$\\
\multirow{4}{*}[1em]{Multi-Layer} & Exponential & $87.9$ & $72.8$ & $72.3$ & $76.4$ & $81.3$ & $61.7$ & $71.9$ & $79.0$ & $75.4$\\
& P-Sigmoid & $87.5$ & $72.3$ & $72.1$ & $76.1$ & $81.0$ & $61.3$ & $71.8$ & $78.6$ & $75.1$\\ & Multi-Step & $88.5$ & $73.9$ & $73.3$ & $76.9$ & $81.9$ & $62.9$ & $72.2$ & $79.8$ & $76.2$\\
\noalign{
  \hrule height 1pt
}
\rowcolor{gray!20}
\multicolumn{11}{c}{\textit{Reduce FLOPs Budget to 80\%, $\sim$2.84 TFLOPs}} \\
Single-Layer & - & $88.3$ & $73.7$ & $73.1$ & $76.9$ & $81.6$ & $62.8$ & $72.0$ & $79.4$ & $75.7$\\
\noalign{
  \hrule height 0.5pt
}
& Linear & $87.7$ & $73.2$ & $72.6$ & $76.3$ & $81.4$ & $62.4$ & $71.3$ & $78.8$ & $75.5$\\
\multirow{4}{*}[1em]{Multi-Layer} & Exponential & $87.4$ & $72.7$ & $72.2$ & $75.9$ & $81.0$ & $62.2$ & $70.9$ & $78.8$ & $75.1$\\
& P-Sigmoid & $87.1$ & $72.3$ & $71.8$ & $74.9$ & $80.2$ & $61.3$ & $69.9$ & $77.7$ & $74.4$\\ & Multi-Step & $88.5$ & $73.9$ & $73.4$ & $77.2$ & $81.9$ & $63.2$ & $72.4$ & $79.7$ & $76.3$\\
\noalign{
  \hrule height 1pt
}
\rowcolor{gray!20}
\multicolumn{11}{c}{\textit{Reduce FLOPs Budget to 70\%, $\sim$2.50 TFLOPs}} \\
Single-Layer & - & $88.0$ & $73.4$ & $72.8$ & $76.7$ & $80.9$ & $62.4$ & $71.1$ & $78.9$ & $75.5$\\
\noalign{
  \hrule height 0.5pt
}
& Linear & $87.4$ & $72.8$ & $72.2$ & $76.3$ & $80.5$ & $62.1$ & $70.8$ & $78.3$ & $75.1$\\
\multirow{4}{*}[1em]{Multi-Layer} & Exponential & $87.2$ & $72.5$ & $72.0$ & $76.1$ & $80.4$ & $61.8$ & $70.6$ & $78.2$ & $74.9$\\
& P-Sigmoid & $86.8$ & $72.3$ & $71.8$ & $75.7$ & $80.1$ & $61.6$ & $70.4$ & $78.0$ & $74.6$\\ & Multi-Step & $87.7$ & $72.9$ & $72.5$ & $76.6$ & $80.8$ & $62.0$ & $70.6$ & $78.7$ & $75.2$\\
\noalign{
  \hrule height 1pt
}
\rowcolor{gray!20}
\multicolumn{11}{c}{\textit{Reduce FLOPs Budget to 60\%, $\sim$2.14 TFLOPs}} \\
Single-Layer & - & $86.0$ & $71.4$ & $70.8$ & $74.7$ & $78.9$ & $60.4$ & $69.1$ & $76.9$ & $73.5$\\
\noalign{
  \hrule height 0.5pt
}
& Linear & $84.5$ & $69.7$ & $68.9$ & $72.7$ & $77.1$ & $58.1$ & $68.1$ & $74.4$ & $71.7$\\
\multirow{4}{*}[1em]{Multi-Layer} & Exponential & $83.2$ & $68.5$ & $67.7$ & $71.4$ & $76.3$ & $56.8$ & $67.5$ & $73.1$ & $70.6$\\
& P-Sigmoid & $82.9$ & $68.1$ & $66.5$ & $70.1$ & $75.3$ & $56.2$ & $66.5$ & $72.4$ & $69.8$\\ & Multi-Step & $86.3$ & $71.7$ & $71.1$ & $74.9$ & $79.2$ & $60.6$ & $69.3$ & $77.3$ & $73.8$\\
\noalign{
  \hrule height 1pt
}
\rowcolor{gray!20}
\multicolumn{11}{c}{\textit{Reduce FLOPs Budget to 50\%, $\sim$1.78 TFLOPs}} \\
Single-Layer & - & $84.8$ & $69.4$ & $67.6$ & $71.2$ & $77.1$ & $58.1$ & $67.8$ & $75.9$ & $71.5$\\
\noalign{
  \hrule height 0.5pt
}
& Linear & $82.6$ & $67.7$ & $65.4$ & $70.9$ & $76.2$ & $57.5$ & $66.5$ & $74.9$ & $70.2$\\
\multirow{4}{*}[1em]{Multi-Layer} & Exponential & $82.2$ & $67.3$ & $65.1$ & $70.4$ & $75.3$ & $57.1$ & $65.5$ & $74.4$ & $69.7$\\
& P-Sigmoid & $81.9$ & $67.0$ & $64.8$ & $69.6$ & $74.7$ & $56.8$ & $65.2$ & $74.1$ & $69.3$\\ & Multi-Step & $84.9$ & $69.5$ & $68.6$ & $71.8$ & $77.9$ & $59.2$ & $68.5$ & $76.7$ & $72.1$\\
\noalign{
  \hrule height 1pt
}
\rowcolor{gray!20}
\multicolumn{11}{c}{\textit{Reduce FLOPs Budget to 40\%, $\sim$1.42 TFLOPs}} \\
Single-Layer & - & $77.6$ & $61.7$ & $62.6$ & $64.4$ & $72.6$ & $54.1$ & $62.5$ & $69.9$ & $65.7$\\
\noalign{
  \hrule height 0.5pt
}
& Linear & $76.2$ & $59.4$ & $60.3$ & $62.6$ & $70.3$ & $52.7$ & $61.3$ & $67.7$ & $63.8$\\
\multirow{4}{*}[1em]{Multi-Layer} & Exponential & $76.1$ & $59.2$ & $59.6$ & $62.3$ & $69.6$ & $52.3$ & $50.7$ & $67.4$ & $62.2$\\
& P-Sigmoid & $73.3$ & $56.3$ & $56.7$ & $60.5$ & $66.9$ & $49.1$ & $47.5$ & $65.7$ & $59.5$\\ & Multi-Step & $77.8$ & $62.1$ & $62.9$ & $64.8$ & $73.3$ & $55.3$ & $62.6$ & $66.1$ & $65.7$\\
\noalign{
  \hrule height 1pt
}
\rowcolor{gray!20}
\multicolumn{11}{c}{\textit{Reduce FLOPs Budget to 30\%, $\sim$1.06 TFLOPs}} \\
Single-Layer & - & $70.2$ & $54.6$ & $55.5$ & $54.5$ & $65.7$ & $46.4$ & $54.4$ & $62.8$ & $58.0$\\
\noalign{
  \hrule height 0.5pt
}
& Linear & $65.5$ & $49.6$ & $50.8$ & $49.3$ & $61.3$ & $41.8$ & $49.4$ & $57.8$ & $53.2$\\
\multirow{4}{*}[1em]{Multi-Layer} & Exponential & $65.3$ & $49.5$ & $50.6$ & $49.2$ & $60.9$ & $41.6$ & $49.3$ & $57.7$ & $53.0$\\
& P-Sigmoid & $60.9$ & $45.3$ & $47.2$ & $44.7$ & $55.9$ & $38.3$ & $45.9$ & $52.4$ & $48.8$ \\ & Multi-Step & $71.3$ & $55.7$ & $56.2$ & $55.6$ & $66.2$ & $47.6$ & $55.6$ & $63.2$ & $58.9$\\
\noalign{
  \hrule height 1pt
}
\rowcolor{gray!20}
\multicolumn{11}{c}{\textit{Reduce FLOPs Budget to 20\%, $\sim$0.72 TFLOPs}} \\
Single-Layer & - & $46.6$ & $43.1$ & $50.1$ & $43.9$ & $51.2$ & $38.9$ & $17.8$ & $48.5$ & $42.5$\\
\noalign{
  \hrule height 0.5pt
}
& Linear & $44.8$ & $41.7$ & $48.2$ & $41.4$ & $48.7$ & $36.4$ & $15.3$ & $46.4$ & $40.4$\\
\multirow{4}{*}[1em]{Multi-Layer} & Exponential & $43.9$ & $40.2$ & $47.2$ & $40.4$ & $47.4$ & $35.8$ & $14.5$ & $45.4$ & $39.4$\\
& P-Sigmoid & $42.1$ & $39.2$ & $46.1$ & $39.7$ & $46.6$ & $34.3$ & $13.8$ & $44.5$ & $38.3$\\ & Multi-Step & $46.7$ & $43.2$ & $50.5$ & $43.2$ & $51.4$ & $38.8$ & $18.6$ & $47.9$ & $42.5$\\
\noalign{
  \hrule height 1pt
}
\rowcolor{gray!20}
\multicolumn{11}{c}{\textit{Reduce FLOPs Budget to 10\%, $\sim$0.36 TFLOPs}} \\
Single-Layer & - & $35.5$ & $31.7$ & $46.9$ & $35.5$ & $40.1$ & $33.2$ & $10.1$ & $36.1$ & $33.6$\\
\noalign{
  \hrule height 0.5pt
}
& Linear & $35.2$ & $31.4$ & $46.7$ & $35.2$ & $39.8$ & $32.9$ & $9.9$ & $35.7$ & $33.6$\\
\multirow{4}{*}[1em]{Multi-Layer} & Exponential & $34.4$ & $30.9$ & $46.4$ & $34.9$ & $39.1$ & $31.9$ & $9.3$ & $35.2$ & $32.8$\\
& P-Sigmoid & $34.6$ & $31.3$ & $46.8$ & $35.2$ & $39.4$ & $32.3$ & $9.8$ & $35.5$ & $33.1$\\
& Multi-Step & $35.4$ & $31.1$ & $46.3$ & $35.3$ & $39.7$ & $32.8$ & $10.0$ & $35.8$ & $33.3$\\
\noalign{
  \hrule height 1pt
}
\end{tabular}}
\vspace{-1em}
\end{table*}

\end{document}